\documentclass{article}

\usepackage[main,final]{neurips_2026}

\usepackage[utf8]{inputenc} % allow utf-8 input
\usepackage[T1]{fontenc}    % use 8-bit T1 fonts
\usepackage{hyperref}       % hyperlinks
\usepackage{url}            % simple URL typesetting
\usepackage{booktabs}       % professional-quality tables
\usepackage{amsfonts}       % blackboard math symbols
\usepackage{nicefrac}       % compact symbols for 1/2, etc.
\usepackage{microtype}      % microtypography
\usepackage{xcolor}         % colors
\usepackage{amsmath}
\usepackage{graphicx}
\usepackage{algorithm}
\usepackage{subcaption}
\usepackage{multirow}
\usepackage{adjustbox}
\usepackage{amssymb,amsthm}
\newtheorem{proposition}{Proposition}
\usepackage{algpseudocode}
\DeclareMathOperator*{\argmax}{arg\,max}
\DeclareMathOperator*{\argmin}{arg\,min}
\newcommand{\model}{\textsc{RAD-TFM}}
\newcommand{\modelfull}{\textsc{Robust and Domain-Adapted Tabular Foundation Models}}
\title{\model: Robust and Domain-Adapted Tabular Foundation Models}

\author{%
  Matthew Peroni \\
  MIT, IBM\\
  \texttt{mperoni@ibm.com}\\
  \And
  Franck Le \\
  IBM \\
  \texttt{fle@us.ibm.com} \\
  \And
  Vadim Sheinin \\
  IBM \\
  \texttt{vadims@us.ibm.com}
}

\begin{document}

\maketitle

\begin{abstract}
The development of tabular foundation models (TFMs) has accelerated in recent years, outperforming traditional ML methods on many tabular benchmarks. A key finding is that TFMs can be pretrained entirely on synthetic datasets, opening opportunities to design data generators that encourage desirable model properties. Prior work has mainly focused on crafting high-quality priors over generators to improve overall pretraining performance. Our key insight is that parameterizing the generator distribution enables an adversarial, distributional robustness perspective: during training, we can adapt the generator to emphasize datasets that are particularly challenging for the model, which we formalize by introducing an optimality gap measure.
% given by the difference between TFM performance and the best achievable performance as estimated by strong baselines such as XGBoost, CatBoost, and Random Forests
Further, we develop a method to align the synthetic data generation with real-world datasets from a given domain, constraining the adversary to generate realistic data.
Together, these algorithms comprise the \modelfull\ (\model) pipeline, a model-agnostic adversarial training framework. Applied to the TabPFN V2 classifier, \model\ improves performance across 6 diverse tabular benchmarks, with up to a 10.8 percentage-point improvement in mean normalized AUC over the strongest baseline, using fewer than 100k additional synthetic training datasets, less than $0.1\%$ of the original pretraining data. Applying the same pipeline to TabICL V2 also improves performance across three major benchmarks, demonstrating the generality of the framework. These results highlight a promising new direction for targeted adversarial continued training of TFMs using synthetic data alone.
\end{abstract}

\section{Introduction}
Recent survey studies have challenged the deep learning (DL) community to improve upon traditional methods for learning from tabular data, such as  boosted trees methods, having found that DL approaches still lagged behind in terms of performance \cite{grinsztajn_why_2022}. In response, substantial progress has been made to close this gap, resulting in a variety of strong deep learning approaches being proposed over the past few years \citep{somepalli_saint_2021, gorishniy_revisiting_2023, hegselmann_tabllm_2023, hollmann_tabpfn_2023, carballo_tabtext_2023, yan_making_2024, link_mitra_2025}. Among these new approaches, tabular foundation models (TFMs) which rely on in-context learning (ICL) \citep{hollmann_tabpfn_2023, qu_tabicl_2025, link_mitra_2025} have emerged as a promising direction for classification and regression tasks with structured datasets. The benefits of this approach are two-fold. In many cases, the performance of the TFMs improves upon strong traditional baselines such as boosted trees \citep{friedman_greedy_2001, erickson_tabarena_2025}. Additionally, this performance is achieved in a zero-shot framework, enabling high-quality predictions on new datasets in milliseconds when GPU-accelerated. Further, as we will explore in this work, since these TFMs are pretrained using synthetic data, there is significant opportunity for improvement as the data generation process itself is expanded and improved.

Pretraining TFMs relies on generating a large amount of diverse synthetic datasets \citep{hollmann_tabpfn_2023}. As we will formally define in the following section, this generation process relies on constructing structural causal models (SCMs) from which datasets can be sampled \citep{pearl_causality_2009, shanmugam_elements_2018}. The structure of these SCMs is implicitly parameterized, giving significant control over the data generation process. All current publicly available, competitive TFMs  \citep{hollmann_tabpfn_2023, qu_tabicl_2025, link_mitra_2025} have been pretrained on datasets generated from a fixed prior distribution over these SCM parameters. However, fixed priors underrepresent certain regions of the parameter space, potentially degrading performance on real-world datasets with similar structure. This contributes to state-of-the-art TFMs still lagging behind tree-based methods on some benchmarks \citep{mcelfresh_when_2024, ye_closer_2025}. In this work, we leverage the significant control provided by the data generation process to frame TFM training from an adversarial robustness perspective \citep{madry_towards_2019, rahimian2019distributionally}. Recent work has explored a narrower version of this idea, focusing on adjusting the weights of a specific class of SCMs in GAN-style training of TFMs \cite{wu_zero-shot_2025}. In this work, we focus on a broad and highly flexible framework for adversarially robust training of TFMs. Going one step further, we ground this adversarial training process, constraining it to generating realistic data, yielding a robust and domain-adapted TFM.

Recent developments in TFMs, such as Real-TabPFN, have demonstrated the benefit of incorporating real-world data into the training process of TFMs \cite{garg2025realtabpfnimprovingtabularfoundation}. However, as we will show in this work, finetuning on real datasets alone is often insufficient due to the lack of large-scale, high-quality tabular datasets, especially for more specific domains such as genomics or physical sciences. In this work, we will train TFMs on synthetic data, but we will constrain the data generation process to produce datasets that are similar to real-world datasets from a given domain. Previous work that aims to align a synthetic data generator with a tabular dataset typically take a GAN-based approach, focusing on constructing a generator to closely mimic a single real dataset \cite{choi2018generatingmultilabeldiscretepatient, Park_2018, xu2019modelingtabulardatausing}. In this work, we will expand beyond this one-to-one relationship, developing an algorithm that efficiently constructs a \emph{distribution over SCM-based tabular dataset generators} that, in aggregate, align with real-world data across a large variety of meta-features capturing data schema, geometry, and learnability.
\begin{figure}[t]
    \centering
    \includegraphics[width=\linewidth]{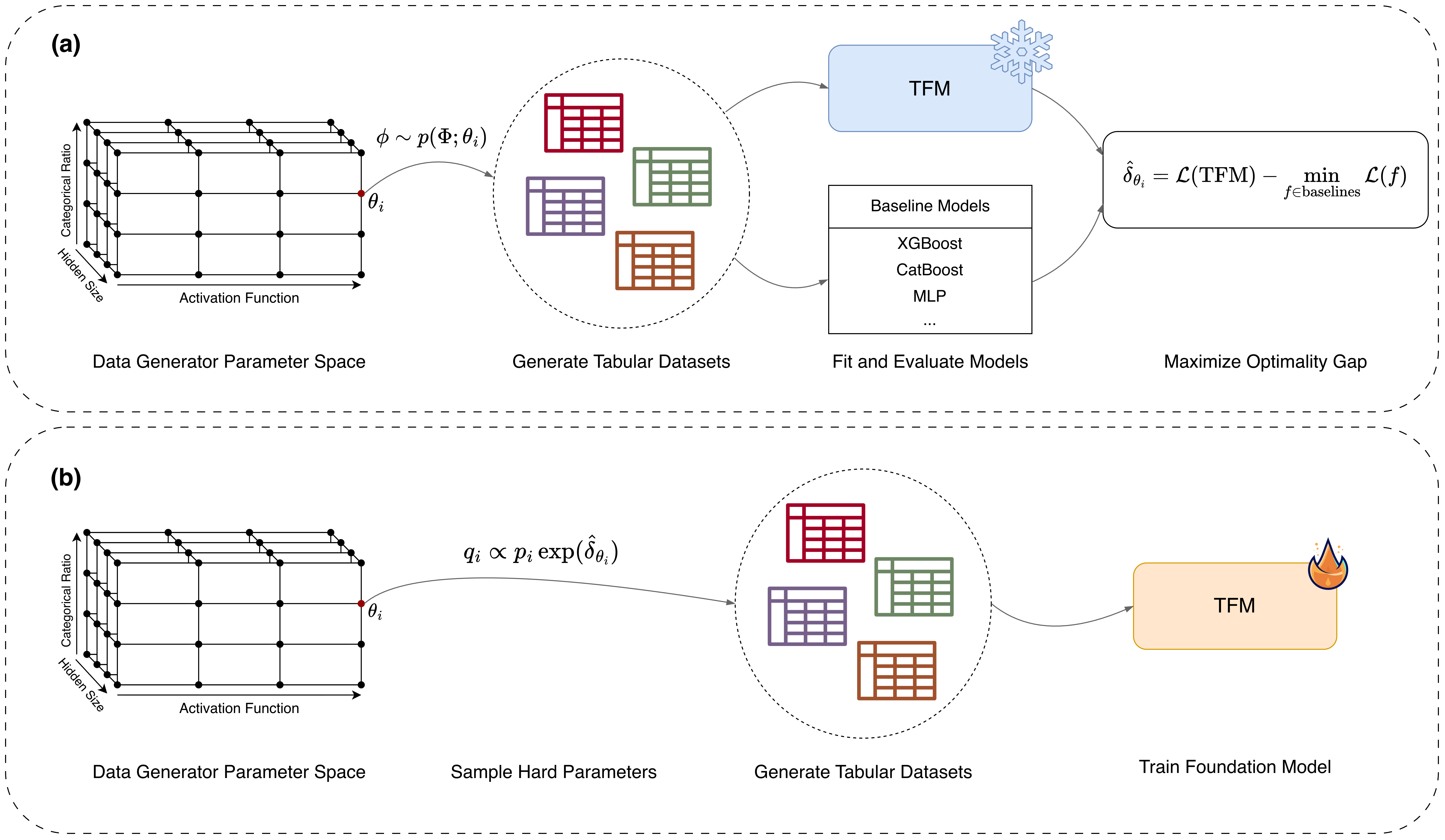}
    \caption{Overview of \model. We solve the core distributionally robust optimization problem by alternating between two stages: \textbf{(a)} searching over the SCM parameter space, we find synthetic data distributions with large estimated optimality gaps, and \textbf{(b)} sampling parameters proportional to their optimality gap, we generate synthetic datasets and continue training the TFM.}
    \label{fig:rtfm}
\end{figure}

Our contributions can be summarized as follows: \textbf{1)} We formalize distributionally-robust adversarial training over the SCM parameter space, allowing the model to adapt to challenging regions (e.g., varying numbers of features, categorical feature ratios, nonlinearities). We introduce an \textit{optimality gap} objective and use it to target regions where the TFM underperforms relative to the best achievable performance. \textbf{2)} We formalize the notion of domain-adapted TFMs in our DRO framework and introduce an efficient algorithm to construct a reference distribution over synthetic data generators that closely matches the real-data distribution for a given domain. \textbf{3)} We propose an efficient, model-agnostic two-stage adversarial training algorithm for TFMs, which we call \modelfull\ (\model). \textbf{4)} We apply \model\ to TabPFN V2 \citep{hollmann_accurate_2025}, showing that with fewer than 100k additional training datasets, less than $0.1\%$ of the original pretraining data, we can significantly improve its ranking on several real-world tabular benchmarks. We further demonstrate that the same training pipeline improves TabICL V2 \citep{qu_tabiclv2_2026}, an architecturally distinct TFM. To organize this work, in Section \ref{sec:problem}, we first define the DRO formulation for robust and domain-adapted training of TFMS. Then, in Section \ref{sec:da}, we introduce our method for finding an approximate reference distribution over data generators that aligns with real-world datasets. With these elements developed, in Section \ref{sec:rtfm}, we define our alternating two-stage algorithm for solving the core DRO problem. Finally, in Section \ref{sec:experiments}, we apply our \model\ pipeline and evaluate it on a variety of tabular benchmarks, both general and application-specific.

\section{Problem Definition}\label{sec:problem}
We begin by defining some notation, and then introduce our theoretical formulation for \model. Let $\Phi$ represent the space of hypothesis functions which define the relationship between features $x$ and outcome $y$. The outcome $y$ can be discrete or continuous, but for this work, we focus on classification tasks. Then a hypothesis $\phi \in \Phi$ is a specific function which we can treat as a data generating mechanism. In the case of SCMs, $\Phi$ is the entire space of SCMs, and $\phi \in \Phi$ is a specific instance of an SCM, from which a dataset $D \sim p(\phi)$ can be generated. In this work, we implement SCMs as randomly initialized multi-layer perceptrons (MLPs), as introduced in \cite{hollmann_tabpfn_2023}. Full details of this process are provided in Appendix \ref{app:scm}. For example, we can sample the number of layers $l \sim p(l)$, the hidden size $h \sim p(h)$, the activation function $a \sim p(a)$, and the ratio of categorical features $r_{cat} \sim p(r_{cat})$, among other dimensions. An SCM is generated by first sampling from the joint distribution over these hyperparameters, each of which is parameterized. For example, we might sample the ratio of categorical features $r_{cat} \sim TruncNorm(\mu_{r_{cat}}, a=0, b=1)$ from the truncated normal distribution with some mean $\mu_{r_{cat}}$ which parametrizes the distribution. Then, assuming we have similar parameterizations for the distribution over each hyperparameter, we define $\theta = (\mu_l, \mu_h, \mu_{r_{cat}}, \dots)$ as the parameterization of the joint distribution over all the hyperparameters. We define $\Theta$ as the space of the parameters, such that $\theta \in \Theta$ represents a specific parameterization. The generators are then sampled from $\phi \sim p(\Phi;\theta)$.

 We now define the original objective for the Prior-Fitted Network (PFN) learning task and provide our adversarially robust formulation. Let $g_{\mathbf{W}}$ be our predictive model parameterized by weight matrix $\mathbf{W}$. As a PFN, $g_{\mathbf{W}}$ takes as input a sequence of labeled training samples and unlabeled test samples, and predicts labels for the test samples. For a given data generator $\phi \in \Phi$, the PFN loss is defined as the cross entropy loss over the test samples,
\begin{equation}
    \mathcal{L}_{PFN}(\mathbf{W}; \phi) =
\mathbb{E}_{(\{(x_{t},y_{t})\} \cup D) \sim p(\phi)}[-\log(g_{\mathbf{W}}(y_{t}|x_{t}, D)]. 
\end{equation}
Since the model will be trained by sampling many data generators from the distribution $\phi \sim p(\Phi ; \theta)$, we can write the original optimization problem from \cite{hollmann_tabpfn_2023} as follows,
\begin{equation}
    \min_{\mathbf{W}} \mathbb{E}_{\phi \sim p(\Phi ; \theta)} [\mathcal{L}_{PFN}(\mathbf{W}; \phi)].
\end{equation}
Since our goal is to learn a foundation model which performs well across all data generation schemes, we want to introduce an adversary who can shift the distribution over data generators to identify the regions over which $g_{\mathbf{W}}$ yields the worst performance. However, there is a key subtlety in this formulation. Maximizing the loss alone could lead the maximizer to select data generators which produce datasets that are hard to learn for any model. Instead, we want to maximize the \textit{optimality gap} between the model's current performance and the best achievable performance. Since we are using the cross-entropy loss, we can frame this optimality gap in terms of the conditional entropy of the data. For any data generator $\phi$, let us denote the features as a random variable $Z_x$ and the target as $Z_y$ such that $(Z_x, Z_y) \sim p(\phi)$. Then the Bayes' optimal predictor is exactly $f^* = p(Z_y | Z_x)$ which achieves the cross-entropy loss $L_{CE}(f^*) = H_{\phi}(Z_y | Z_x)$, the conditional entropy of the target $Z_y$ given the features $Z_x$. It follows that for any fixed data generator, $\mathcal{L}_{PFN} \geq H(Z_y | Z_x)$, and $\mathcal{L}_{PFN} - H(Z_y | Z_x) \geq 0$ is the optimality gap. We can then state the adversarially robust optimization problem in terms of maximizing the optimality gap,
\begin{equation}\label{eq:main-objective}
    \min_{\mathbf{W}}\max_{\theta \in \Theta} \mathbb{E}_{\phi \sim p(\Phi ; \theta)} [\mathcal{L}_{PFN}(\mathbf{W}; \phi) - H_{\phi}(Z_y | Z_x)].
\end{equation}
For brevity, we denote the objective function as $\delta_\theta(\mathbf{W}) = \mathbb{E}_{\phi \sim p(\Phi ; \theta)} [\mathcal{L}_{PFN}(\mathbf{W}; \phi) - H_{\phi}(Z_y | Z_x)]$. While Equation \ref{eq:main-objective} is the precise theoretical optimization problem we wish to solve, in reality, estimating the conditional entropy $H_{\phi}(Z_y | Z_x)$ or evaluating it analytically is often infeasible. Therefore, rather than estimate $H_{\phi}(Z_y | Z_x)$ directly, in this work, we replace this term with the minimum cross-entropy loss over a collection of baseline estimators $f_1, \dots, f_e$, such as XGBoost, Catboost, and Random Forests \cite{chen_xgboost_2016,dorogush_catboost_2018, breiman_random_2001}, which are fit to each generated dataset. Since $\min_{i \in [e]}\mathcal{L}_{PFN}(f_i;\phi) \geq H(Z_y | Z_x)$, it follows that $\mathbb{E}_{\phi \sim p(\Phi;\theta)}[\mathcal{L}_{PFN}(\mathbf{W}; \phi) - \min_{i \in [e]} \mathcal{L}_{PFN}(f_i; \phi)] \leq \delta_\theta(\mathbf{W})$, providing us with a lower bound on the optimality gap. We then modify our original objective function, defining 
\begin{equation}
    \widehat\delta_\theta(\mathbf{W}) = \mathbb{E}_{\phi \sim p(\Phi;\theta)}[\mathcal{L}_{PFN}(\mathbf{W}; \phi) - \min_{i \in [e]} \mathcal{L}_{PFN}(f_i; \phi)].
\end{equation}
Since $\widehat \delta_\theta(\mathbf{W}) \leq \delta_\theta(\mathbf{W})$, it follows that $\widehat\delta_\theta(\mathbf{W}) > 0$ implies $\delta_\theta(\mathbf{W}) > 0$, such that for any parameter estimated to have a positive optimality gap, the model $g_{\mathbf{W}}$ does indeed have room for improvement. Further, this lower bound can always be improved by adding more baseline estimators with varying strengths and inductive biases. In practice, maximizing over a single parameterization could lead to the model overfitting to a particular region of the parameter space. Further, for any given domain or application, we want to ensure that the generated synthetic datasets do not stray too far from the types of datasets observed in that domain. We therefore frame this problem from a distributionally robust optimization (DRO) perspective by constraining the actions of the maximizer \cite{rahimian2019distributionally}. Rather than a specific parameterization $\theta$, we allow the maximizer to select a distribution $Q \in \Delta_{\Theta}$ over parameters, where $\Delta_{\Theta}$ is the simplex over $\Theta$. We assume $\Theta$ is discrete, which is reasonable in practice since the parameters are either categorical or can be easily discretized (e.g. $\mu_{r_{cat}} \in \{0, 0.1, \dots, 1\}$). To constrain the maximizer, we will assume access to a reference distribution $P \in \Delta_\Theta$ that aligns with the data from a given domain. In Section \ref{sec:da}, we will introduce our methodology for finding such a reference distribution given a corpus of real-world datasets from a specific domain. Then, with the reference distribution $P$, we can constrain the KL-divergence between $Q$ and $P$, leading to the following optimization problem,
\begin{equation}\label{eq:dro-objective}
    \min_{\mathbf{W}}\max_{\substack{
        Q \in \Delta_{\Theta} \\ D_{KL}(Q || P) \leq \rho}} \mathbb{E}_{\theta \sim Q}[\hat\delta_\theta(\mathbf{W})].
\end{equation}
As we prove in Appendix \ref{app:proofs}, the optimal distribution for the maximizer in Equation \ref{eq:dro-objective} is a softmax distribution. Specifically, denoting $Q^* = (q_1^*,\dots,q_{|\Theta|}^*)$ as the optimal distribution and $P=(p_1,\dots p_{|\Theta|})$ as the reference distribution, for any $i \in |\Theta|$, $q_i^* \propto p_i\exp(\eta \cdot \delta_{\theta_i}(\mathbf{W}))$. The temperature $\eta  = \eta(\rho, \delta_\theta(\mathbf{W}), P)$ is uniquely determined by the maximum KL-divergence, the optimality gaps, and the reference distribution. The DRO problem can then be rewritten as
\begin{equation}\label{eq:general-robust-opt}
    \min_{\mathbf{W}} \sum_{i=1}^{|\Theta|} 
    \frac{p_i\exp(\eta \cdot \widehat\delta_{\theta_i}(\mathbf{W}))}{\sum_{i \in |\Theta|}p_i\exp(\eta \cdot \widehat\delta_{\theta_i}(\mathbf{W}))} \cdot
    \widehat\delta_{\theta_i}(\mathbf{W}).
\end{equation}
 In the following sections, we introduce our \model\ approach to solving Equation \ref{eq:general-robust-opt}. In Section \ref{sec:da}, we develop our key contribution to domain adaptation for TFMs, learning the reference distribution $P$ over data generator parameters to align the generated data with real-world data. Then in Section \ref{sec:rtfm}, we incorporate the learned reference distribution into our DRO formulation and develop our method for solving this optimization problem using an iterative, two-stage approach.

\section{Domain-Adapted Tabular Foundation Models}\label{sec:da}
\begin{figure}[t]
    \centering
    \includegraphics[width=\linewidth]{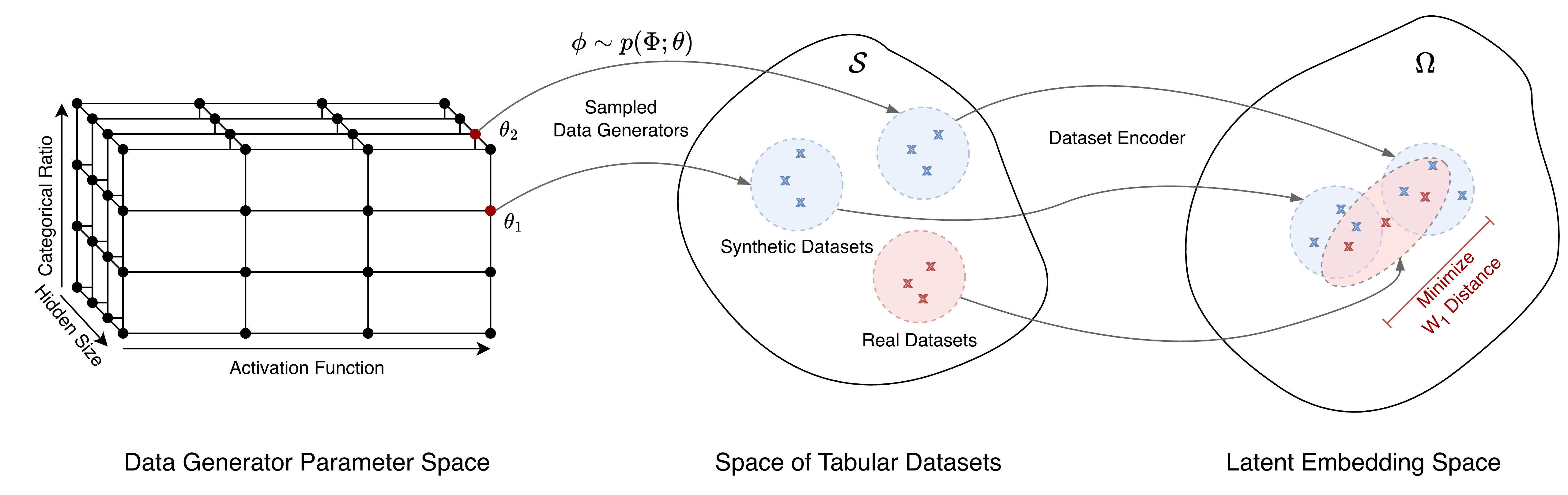}
    \caption{We perform domain adaptation for TFMs (e.g. healthcare, finance) by learning a mixture distribution over SCM generators that covers the distribution of real-world datasets from the given domain. We use a collection of meta-features capturing \emph{dataset schema, geometry, and learnability} to encode and measure distance between datasets.}
    \label{fig:da-illustrate}
\end{figure}
In our \model\ methodology, the key to adapting a TFM to a specific domain is successfully constructing a reference distribution $P \in \Delta_{\Theta}$ that connects a corpus of real-world datasets to a corresponding distribution over dataset generators. We will then use this reference distribution in our DRO formulation given in Equation \ref{eq:dro-objective} to adversarially generate datasets that challenge the TFM while remaining similar to the real-world datasets observed in the given domain. However, tabular datasets are structurally heterogeneous, with varying numbers and types of features, patterns, and targets. This motivates the need for a shared latent space for all tabular datasets that encodes semantic meaning through which we can project and compare datasets. To accomplish this, we rely on meta-feature extraction, a well-studied topic in meta-learning \cite{brazdil2009metalearning}, to efficiently summarize any given tabular dataset into a fixed-size vector of statistical and information-theoretic properties. An illustration of our domain-adaptation scheme is given in Figure \ref{fig:da-illustrate}. Full details are given in Appendix \ref{app:da}.
% Meta-feature stuff goes here!
% One promising method for encoding diverse tabular datasets is to use the latent feature space learned by the TFMs themselves, which has been shown to encode semantically meaningful relationships between datasets \cite{ye_closer_2025, hollmann2025accurate}. In this work, we will use pretrained TFMs as tabular dataset encoders to efficiently represent datasets from any domain in a shared latent space, without requiring any additional training. 

Following the definition of meta-features from \cite{pinto2016towards} and \cite{rivolli2019characterizingclassificationdatasetsstudy}, for a given dataset $(\mathbf{X}, \mathbf{y}) \in \mathcal{S}$, where $S = \cup_{n \leq N, d \leq D} \mathbb{R}^{n \times d}$ represents the space of tabular datasets up to some maximum size $N \times D$, we can construct a \emph{characterization measure} $m: \mathcal{S} \rightarrow \cup_d\mathbb{R}^d$, which produces a variable number of values depending on the shape of the dataset. For example, $m$ might extract the skewness of each numeric feature in the dataset separately. We can then introduce a \emph{summarization function} $\sigma: \cup_d \mathbb{R}^d \rightarrow \mathbb{R}^k$ which condenses the characterization measure (e.g. by taking averages) into a fixed-size summarization vector of dimension $k$. Combining these functions, we define the tabular dataset encoder as $f((\mathbf{X}, \mathbf{y})) = \sigma(m((\mathbf{X}, \mathbf{y})))$, projecting any tabular dataset into a semantically meaningful latent space. We provide details of the exact meta-features used in this work in Appendix \ref{app:da}. Using this encoder, we can reason about the relationship between real datasets and synthetic datasets generated from the class of SCMs.

Suppose we have a corpus $D_r = \{(\mathbf{X}_i, \mathbf{y}_i)\}_{i=1}^{n_r} \in \mathcal{S}^{n_r}$ of real datasets from a given domain. At a high level, our algorithm works as follows: (1) \emph{encode} each real dataset in the corpus as $f((\mathbf{X}_i, \mathbf{y}_i)) = \mathbf{e}_i$ using our meta-feature extractor; (2) use Bayesian Optimization (BO) to find distributions over synthetic data generators such that the embeddings of the synthetic datasets \emph{provide partial coverage} of the real dataset embeddings; (3) Aggregate the results to form a \emph{global distribution over data generators} that minimizes the $W_1$-distance from the distribution of real datasets in the latent space. By creating a mixture over synthetic data distributions that each provide partial coverage of the real datasets, we can construct a reference distribution $P$ that aligns with an entire corpus of real datasets.

We now formalize the key steps in our DA algorithm where we align the distributions of real and synthetic datasets. We provide additional details in Appendix \ref{app:da}. For each $\theta \in \Theta$ we have a distribution $\phi \sim p(\Phi;\theta)$ over data generators and, using the dataset encoder, a distribution $\mathbf{e}(\theta) \sim p(f(\phi); \theta)$ over dataset embeddings, where $f(\phi)$ is short for the embedding of $(\mathbf{X}, y) \sim p(\phi)$, the dataset sampled from the specific generator. If we assume $|\Theta| = M$ and denote $\Theta = \{\theta_1,\dots, \theta_M\}$, we can define $Q(\theta_i) = p(f(\phi); \theta_i)$ as the distribution over datasets in the latent space induced by $\theta_i$. Then, given that the embeddings of the corpus of real datasets is sampled from some distribution $P_r$, our goal is to solve the following optimization problem,
\begin{equation}\label{eq:main-w1}
    P^* \in \argmin_{P \in \Delta_\Theta} W_1\bigg( P_r, \sum_{i=1}^M p_i \cdot Q(\theta_i) \bigg ),
\end{equation}
finding the mixture over these parameterized generator distributions that minimizes the 1-Wasserstein distance from the real data distribution. Solving this optimization problem recovers the reference distribution $P^*$ with the best possible coverage of the underlying domain-specific data distribution. In practice, we instead optimize the entropy-regularized optimal transport distance, which can be efficiently solved via the Sinkhorn Algorithm \cite{cuturi2013sinkhorn}.
% In practice, we decompose this problem by first clustering the embeddings of the real datasets and solving Equation \ref{eq:main-w1} with cluster-specific distributions $P_r^c$. 
Since $M$ is large for any reasonable parameter search space, we approximate this search using Bayesian Optimization \cite{frazier2018tutorial} with a surrogate objective designed to reward partial coverage of $P_r$. The resulting candidate distributions $Q(\theta_i)$ are then combined by solving the mixture optimization problem in Equation \ref{eq:main-w1}, which has the side effect of pruning redundant and unhelpful generator distributions. The details of our method are given in Algorithm \ref{alg:da}.

\section{Robust Tabular Foundation Models}\label{sec:rtfm}
\begin{figure}[t]
    \centering
    \begin{subfigure}[t]{0.32\linewidth}
        \centering
        \includegraphics[width=\linewidth]{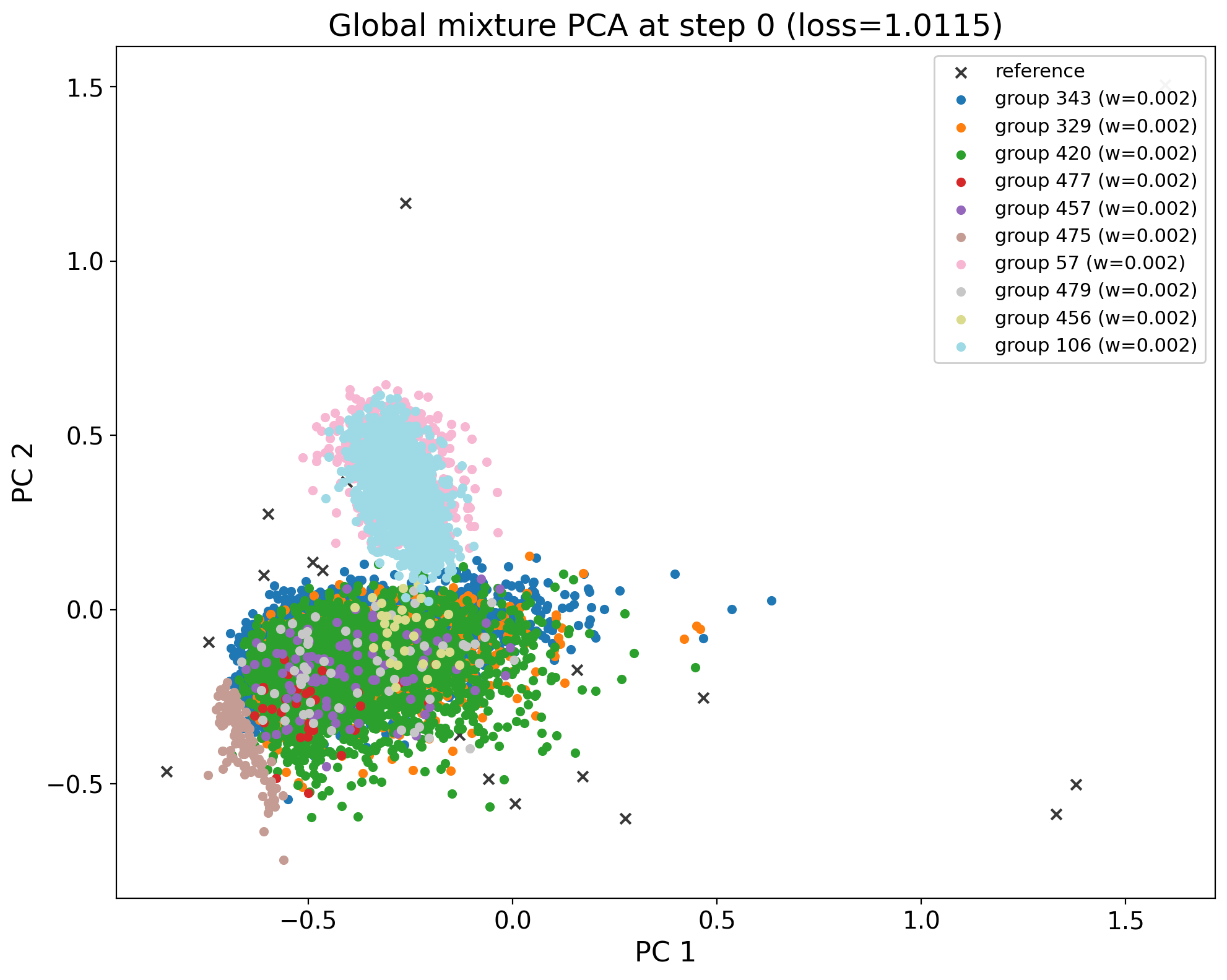}
        % \caption{First subfigure.}
        \label{fig:example-a}
    \end{subfigure}
    \hfill
    \begin{subfigure}[t]{0.32\linewidth}
        \centering
        \includegraphics[width=\linewidth]{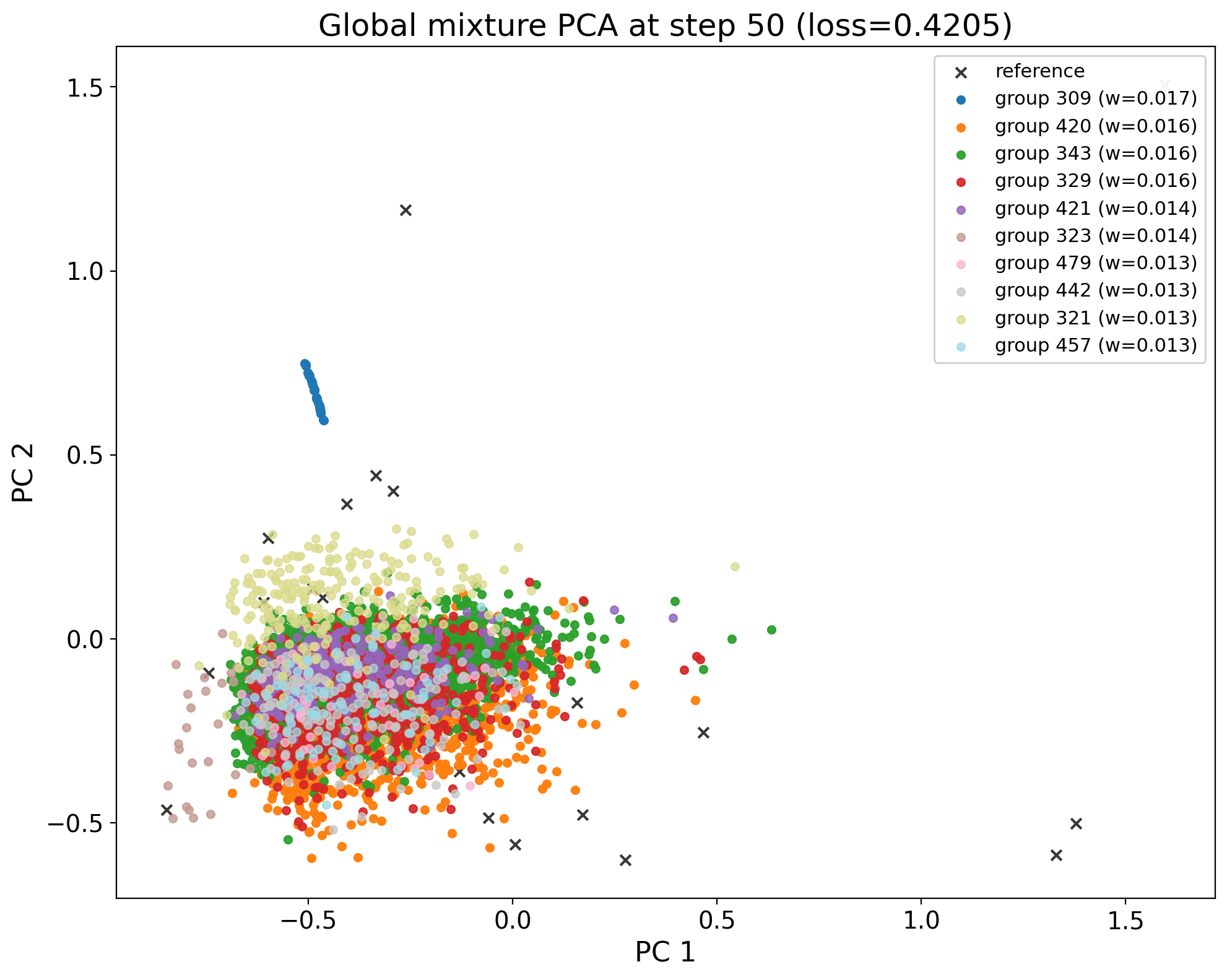}
        % \caption{Second subfigure.}
        \label{fig:example-b}
    \end{subfigure}
    \hfill
    \begin{subfigure}[t]{0.32\linewidth}
        \centering
        \includegraphics[width=\linewidth]{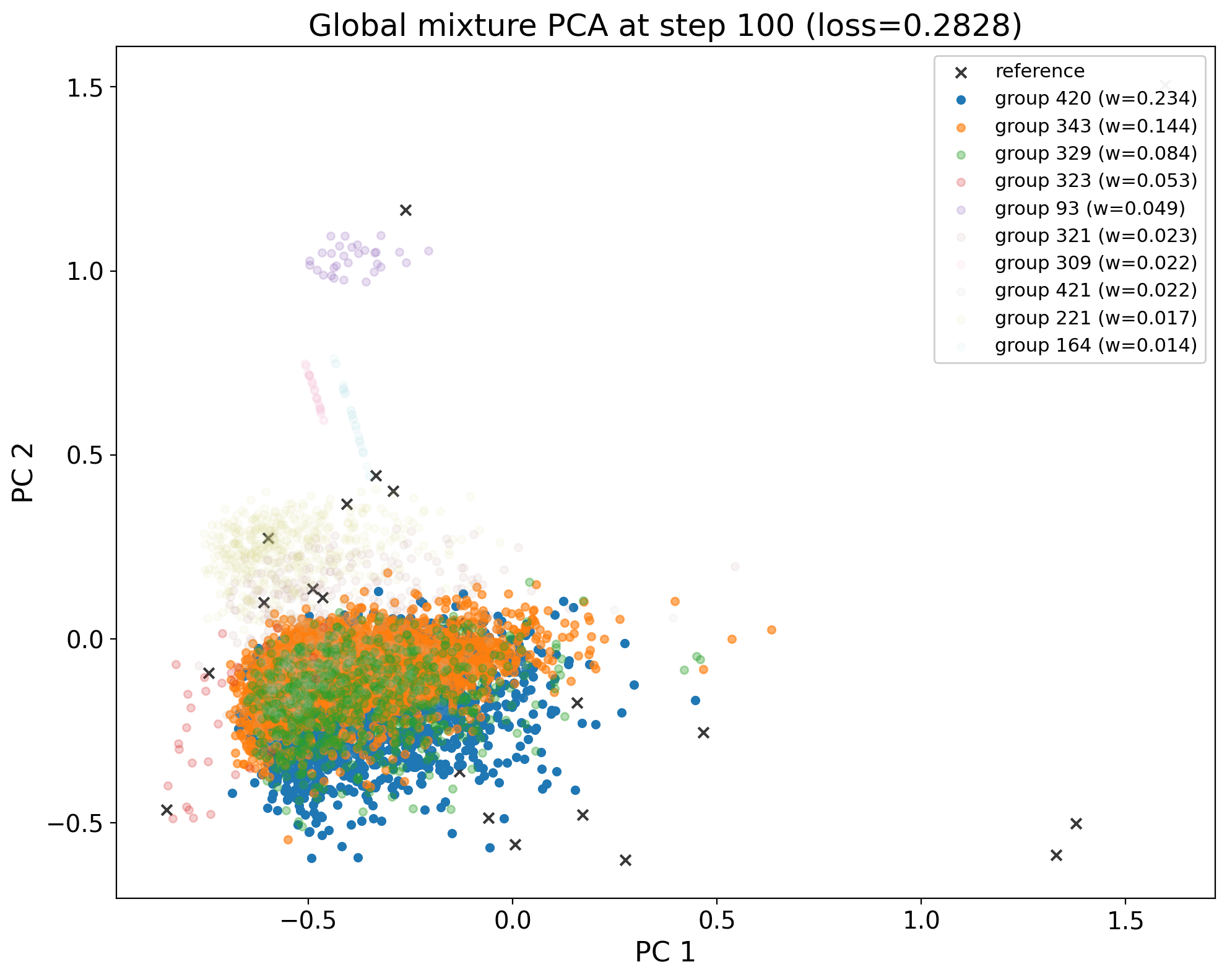}
        % \caption{Third subfigure.}
        \label{fig:example-c}
    \end{subfigure}

    \vspace{-0.5em}
    \caption{An example of our domain-adaptation methodology applied to the engineering datasets from OpenML \cite{bischl_openml_2021}. Each color corresponds to a different parameterization of the synthetic data generator distribution. The black marks represent the embeddings of the real datasets. We plot the first two principal components of the meta-features extracted for the real datasets.}
    \label{fig:three-subfigures}
    \vspace{-0.5em}
\end{figure}
In this section, we give an overview of our two-stage optimization algorithm, \modelfull\ (\model), which is visualized in Figure \ref{fig:rtfm}. The pseudocode and additional details are provided in Appendix \ref{app:rtfm}. We first decompose the problem naturally into a maximization stage (parameter search) and a minimization stage (model training). We assume access to a pretrained TFM, $g_{\mathbf{W}}$, with weights $\mathbf{W}$ which will be used as the input model for our \model\ algorithm. While not absolutely necessary, starting with a pretrained model gives us a warm-start for the model weights. In addition, we specify a fixed set of baseline models $f_1,\dots,f_e$ which will be used to estimate the optimality gap. We begin with the maximization stage, during which we freeze $g_{\mathbf{W}}$ to maximize the optimality gap in its current state. Since the cardinality of the parameter space is large in practice, we use a black-box optimization algorithm \cite{akiba_optuna_2019, watanabe_tree-structured_2023, bergstra_algorithms_2011} to efficiently search the space for parameters with large optimality gaps. 

Using the black-box optimization algorithm, parameters $\theta_i$ are proposed in sequence. For each $\theta_i$ proposed, we estimate the optimality gap by sampling a fixed number, $n_{ds}$, of generators $\phi_1,\dots, \phi_{n_{ds}} \sim p(\Phi;\theta_i)$ and datasets $d_j\sim p(\phi_j)$ and compute $\widehat{\delta}_{\theta_i} = \frac{1}{n_{ds}}\sum_{j=1}^{n_{ds}}\mathcal{L}_{PFN}(\mathbf{W}; d_j) - \min_{k \in [e]}\mathcal{L}_{PFN}(f_k, d_j)$. Then for each proposed $\theta_i$, we must fit $n_{ds} \cdot e$ baseline estimators. However, each pair of dataset and baseline estimator can be fit independently, such that this step can be trivially parallelized, where each baseline estimator is fit to each dataset in parallel. In practice, we found that for $n_{ds} = 20$ and $e = 7$, the estimated optimality gap $\widehat{\delta}_{\theta_i}$ could be computed in a matter of seconds when parallelized, given sufficient CPU cores (we used $n_{cores} = n_{ds} \cdot e$). We repeat this process for a fixed number of trials, $n_{trials}$, with the underlying optimization algorithm proposing new parameters $\theta_i$ based on the observed estimated optimality gaps. At the end of this stage, we have a collection $\{(\theta_i, \delta_{\theta_i})\}_{i=1}^{n_{trials}}$ of parameters and associated estimated optimality gaps.

In the minimization stage, our goal is to train $g_{\mathbf{W}}$ on the regions of the parameter space with large optimality gaps. Following our derivation in Equation \ref{eq:general-robust-opt}, we define a distribution $Q$ over $\Theta$ such that $q_i \propto p_i\exp(\eta \cdot \delta_{i}(\mathbf{W}))$.  The function \(\eta \mapsto D_{KL}(Q(\eta) || P)\) is strictly increasing on $(0, \infty)$, so we can quickly find $\eta$ given a fixed $\rho$ via one-dimensional root finding (e.g., bisection). Using the observations from the maximization stage, our optimization problem becomes
\begin{equation}\label{eq:general-robust-opt-trials}
    \min_{\mathbf{W}} \sum_{i=1}^{n_{trials}} 
    \frac{p_i\exp(\eta \cdot \widehat\delta_{\theta_i}(\mathbf{W}))}{\sum_{i \in [n_{trials}]}p_i\exp(\eta \cdot \widehat\delta_{\theta_i}(\mathbf{W}))} \cdot
    \widehat\delta_{\theta_i}(\mathbf{W}).
\end{equation}
% Rather than taking a sum over all $|\mathcal{P}|$ possible parameterizations as in Equation \ref{eq:general-robust-opt}, we take the sum over the $n_{trials}$ parameters that were proposed during the maximization stage. 
Then, to train the TFM, we generate each dataset within a batch by sampling $\theta_i \sim Q$ and generator $\phi \sim p(\Phi; \theta_i)$. This sampling and dataset generation process is repeated for each batch during model training such that, in expectation, the loss is weighted according to Equation \ref{eq:general-robust-opt-trials}. After training $g_{\mathbf{W}}$ for $n_{iter}$ iterations, we freeze the model return to the maximization stage, using the updated model weights to generate the next set of estimated optimality gaps. This max-min iteration is repeated until the objective in Equation \ref{eq:general-robust-opt-trials} converges. In addition to the baseline estimators $f_1,\dots,f_e$, after a fixed number of max-min iterations, we incorporate the TFM with its original weights as an additional baseline model. This can help mitigate any unlearning that has occurred, encouraging the model to both improve upon strong baselines as well as its own performance at the start of training.

\section{Experiments}\label{sec:experiments}
\textbf{Experiment Setup.} We evaluate our \model\ algorithm on broad benchmark classification datasets from TabArena \cite{erickson_tabarena_2025}, TabPertNet \cite{ye_towards_2024}, and PMLB \cite{olson2017pmlb}, as well as three domain-specific benchmarks derived from datasets available on OpenML \cite{bischl_openml_2021}. We use the open-source model weights from TabPFN V2 \cite{hollmann_accurate_2025} to initialize our TFM, and further apply the same pipeline without architectural modifications to the TabICL V2 classifier \cite{qu_tabiclv2_2026}. We provide a full description of the experiment setup and results in Appendix \ref{app:experiments}. For the \model\ algorithm, we use $n_{trials}=100$ and $n_{ds}=20$ datasets to estimate the optimality gap for each trial. During the minimization stage, we use a learning rate of $lr=1e\text{-}5$ and a batch size of $b=64$. We set $\rho = 0.1\cdot \rho_{\text{max}}$, where $\rho_{\text{max}}$ is the maximum KL-divergence given the learned reference distribution $P$. Depending on the task, we train on 1,000--3,000 generated datasets per epoch and run the full max-min loop for $e\in[5,30]$ epochs, using at most 90k generated training datasets in any run. For baseline models, we use Random Forest, CatBoost, XGBoost, Logistic Regression, and MLPs. For the primary TabPFN benchmark tables, we also finetune TabPFN-V2 on each evaluation dataset for $10$ epochs and report held-out performance as TabPFN (FT). In a separate compute-controlled comparison, we match fine-tuning to \model\ by training-dataset exposure, batch size, and number of optimizer updates, evaluating checkpoints every 500 training datasets (Appendix \ref{app:compute-matched}). We evaluate the conventional baseline algorithms using their respective default settings, and report TFM results without ensembling. All experiments were run on a single node with one A100 GPU; further details of the compute environment and SCM parameter space are provided in Appendices \ref{app:experiments} and \ref{app:scm}. Following previous work, we report normalized ROC-AUC as our primary metric, which scales raw ROC-AUC scores to [0,1], with the worst-performing model receiving a score of 0, and the best a score of 1 \cite{hollmann2025accurate,garg2025realtabpfnimprovingtabularfoundation}. This provides consistency when comparing performance across datasets, and measures how close each model comes to the best possible model on average. For multi-class tasks, we use the One-versus-One ROC-AUC score. We also report the mean rank and number of rank-1 wins for each model, based on the normalized AUC scores.

\begin{table*}[t]
\centering
\caption{Experiment results across three tabular dataset benchmarks. For consistency, we report normalized AUC, with values scaled to [0,1] for each dataset \cite{mcelfresh_when_2024,hollmann_accurate_2025}. For rank-1 wins, we do not count ties, only outright wins.}
\setlength{\tabcolsep}{2pt} % reduce horizontal padding
% \small % smaller font size
\begin{adjustbox}{max width=1\textwidth} % scale table if still too wide
\begin{tabular}{llcccccccc}
\toprule
 \textbf{Benchmark} &\textbf{Metric} & \textbf{LR} & \textbf{MLP} & \textbf{RF} & \textbf{CatBoost} & \textbf{XGB} & \textbf{TabPFN\textsubscript{n.e.}} & \textbf{TabPFN\textsubscript{n.e.}} (\textbf{FT}) & \textbf{TabPFN\textsubscript{n.e.} (\model)} \\
\midrule
\multirow{3}{*}{\textbf{TabPertNet}\cite{ye_towards_2024}} 
& Mean rank AUC & 5.4 & 6.0 & 4.6 & 4.2 & 5.2 & 3.7 & 3.9 & \textbf{3.1} \\
& Mean Norm. AUC  &$0.4684$& $0.3627$ & $0.6646$ & $0.7404$ & $0.5743$ & $0.7306$ & $0.7299$ & $\mathbf{0.8486}^*$ \\
& Rank-1 Wins & 6 & 5 & 7 & 6 & 7 & 9 & 2 & \textbf{13}\\
\midrule
\multirow{3}{*}{\textbf{TabArena}\cite{erickson_tabarena_2025}} 
& Mean rank AUC & 5.7 & 7.2 & 5.5 & 4.0 & 5.2 & 3.6 & 2.7 & \textbf{2.1} \\
& Mean Norm. AUC  &$0.5070$& $0.2513$ & $0.6870$ & $0.8214$ & $0.7293$ & $0.8829$ & $0.9215$ & $\mathbf{0.9421}^*$ \\
& Rank-1 Wins & 2 & 0 & 0 & 1 & 3 & 1 & 4 & \textbf{9}\\
\midrule
\multirow{3}{*}{\textbf{PMLB}\cite{olson2017pmlb}} 
& Mean rank AUC & 5.8 & 7.0 & 4.5 & 4.3 & 4.7 & 3.3 & 3.3 & \textbf{3.1} \\
& Mean Norm. AUC  &$0.4497$& $0.2782$ & $0.7580$ & $0.7798$ & $0.7036$ & $0.8889$ & $0.8894$ & $\mathbf{0.9070}$ \\
& Rank-1 Wins & 8 & 2 & 10 & 8 & 7 & 9 & 4 & \textbf{16}\\
\bottomrule
\end{tabular}
\end{adjustbox}
\label{table:main}
\end{table*}
\begin{table*}[b]
\centering
\caption{Experiment results across three domain-specific tabular dataset benchmarks from OpenML \cite{bischl_openml_2021}. For consistency, we report normalized AUC, with values scaled to [0,1] for each dataset \cite{mcelfresh_when_2024,hollmann_accurate_2025}. For rank-1 wins, we do not count ties, only outright wins.}
\setlength{\tabcolsep}{2pt} % reduce horizontal padding
% \small % smaller font size
\begin{adjustbox}{max width=1\textwidth} % scale table if still too wide
\begin{tabular}{llcccccccc}
\toprule
 \textbf{Benchmark} &\textbf{Metric} & \textbf{LR} & \textbf{MLP} & \textbf{RF} & \textbf{CatBoost} & \textbf{XGB} & \textbf{TabPFN\textsubscript{n.e.}} & \textbf{TabPFN\textsubscript{n.e.}} (\textbf{FT}) & \textbf{TabPFN\textsubscript{n.e.} (\model)} \\
\midrule
\multirow{3}{*}{\textbf{Engineering}} 
& Mean rank AUC & 6.9 & 7.3 & 4.3 & 3.4 & 4.9 & 3.3 & 3.4 & \textbf{2.6} \\
& Mean Norm. AUC  &$0.2944$&$0.3951$& $0.8405$ & $0.8952$ & $0.7805$ & $0.9004$ & $0.8998$ & $\mathbf{0.9313}^*$ \\
& Rank-1 Wins & 0 & 0 & 1 & 4 & 2 & 0 & 0 & \textbf{7}\\
\midrule
\multirow{3}{*}{\textbf{Genomics}} 
& Mean rank AUC & 7.2 & 5.9 & 4.8 & 3.4 & 4.9 & 3.5 & 3.3 & \textbf{2.7} \\
& Mean Norm. AUC  &$0.1481$&$0.4309$& $0.7439$ & $0.8611$ & $0.6979$ & $0.8801$ & $0.8805$ & $\mathbf{0.9088}$ \\
& Rank-1 Wins & 0 & 0 & 1 & 0 & 0 & 0 & 0 & \textbf{2}\\
\midrule
\multirow{3}{*}{\textbf{Healthcare}} 
& Mean rank AUC & 5.3 & 6.9 & 4.5 & 3.5 & 5.0 & 3.8 & 3.8 & \textbf{3.2} \\
& Mean Norm. AUC  &$0.5831$&$0.2050$& $0.7245$ & $0.8446$ & $0.6719$&$0.8696$ & $0.8699$  & $\mathbf{0.8974}$ \\
& Rank-1 Wins & 2 & 0 & 2 & 2 & 4 & 1 & 0 & \textbf{7}\\
\bottomrule
\end{tabular}
\end{adjustbox}
\label{table:main-apps}
\end{table*}
In Table \ref{table:main}, we report the main results from our experiments on the TabArena ($N=21$), TabPertNet ($N=70$), and PMLB ($N=91$) benchmarks. Across our three primary metrics, we observe that \model\ improves performance over all baselines, including the original TabPFN-V2 and the finetuned versions. To measure the significance of our results, we first perform the Friedman test for repeated samples on the median normalized AUC scores \cite{demvsar2006statistical, benavoli2016should}. For the TabPertNet datasets, the Friedman test yielded $p=3.5 \times 10^{-15}$, for TabArena, the test yielded $p=4.3\times10^{-13}$, and for PMLB, the test yielded $p=5.2\times10^{-43}$. Having established significance for the benchmark performance varying by model class, we performed the Wilcoxon signed-rank test pair-wise, comparing the RTFM version of TabPFN to the other models \cite{conover1999practical}. In Tables \ref{table:main} and \ref{table:main-apps}, we add an asterisk to indicate results for which the all Wilcoxon signed-rank tests between \model\ and the alternative models achieves $p < 0.05$. For the main benchmarks reported in Table \ref{table:main}, we see that all results are significant for \model\ for TabPertNet and TabArena. Further, for PMLB, the largest p-value obtained was $p=0.06$, between \model\ and TabPFN-V2 (FT), with $p <0.05$ for all other pair-wise tests. Our results are therefore almost strictly statistically significant, with \model\ achieving the best performance for each metric
across a variety of challenging tabular benchmarks. 

In Table \ref{table:main-apps}, we report results from experiments applying \model\ to domain-specific benchmarks in Engineering ($N=20$), Genomics ($N=11$), and Healthcare ($N=31$), derived from OpenML. To construct these benchmarks, we aggregate all binary and multi-task classification datasets on OpenML matching a given tag (e.g. Engineering), then filter to datasets with at least $n=200$ samples, no more than $n=10k$ samples, and at most $d=500$ features. These benchmarks can be highly domain-specific, and differ significantly from common benchmarks such as TabArena. Similar to the other benchmarks, applying \model\ improves the overall performance of TabPFN-V2 across domains, improving mean rank, normalized AUC, and outright wins. Additionally, for these more specific domains, finetuning TabPFN does not exhibit significant benefits. Across the three domain-specific benchmarks, \model\ achieves 16 rank-1 wins, compared with only one for the original TabPFN and none for the fine-tuned model. The Friedman test returned $p=1.3\times 10^{-11}$, $p=1.3\times 10^{-5}$, and $p=4.1\times 10^{-14}$ for Healthcare, Genomics, and Engineering respectively. Although our approach moves TabPFN's performance in the right direction for these domains, the Wilcoxon signed-rank tests indicate that \model\ only has statistically significant ($p < 0.05$) improvement in normalized AUC for the Engineering benchmark. This is due in part to the small Genomics benchmark ($N=11$). The Healthcare benchmark is particularly challenging, indicating that it may be necessary to expand either the SCM parameter space or the meta-features used to represent the datasets to enable \model\ to properly target this domain. Overall, these results indicate the potential for \model\ to go beyond finetuning TFMs for domain-specific applications, noticeably improving model performance across three diverse domains.

\begin{table*}[t]
\centering
\caption{Comparison of fine-tuning (FT) and \model\ on the TabPFN V2 classifier across three benchmarks. Wins count outright wins among the three TabPFN variants; AUC change reports the relative percentage change in mean raw AUC from the frozen TabPFN. A leap occurs when an adapted variant outperforms at least one conventional baseline that the frozen TabPFN did not outperform.}
\setlength{\tabcolsep}{2pt} % reduce horizontal padding
\small % smaller font size
\begin{adjustbox}{max width=1\textwidth} % scale table if still too wide
\begin{tabular}{llccc}
\toprule
 \textbf{Benchmark} &\textbf{Metric} &  \textbf{TabPFN\textsubscript{n.e.}} & \textbf{TabPFN\textsubscript{n.e.}} (\textbf{FT}) & \textbf{TabPFN\textsubscript{n.e.} (\model)} \\
\midrule
\multirow{4}{*}{\textbf{TabPertNet}\cite{ye_towards_2024}} 
& \#Wins & 11 & 5 & \textbf{37} \\
& AUC Change (\%)  & - & $-0.001$ & $\mathbf{2.061}$ \\
& \#Leaps & - & 1 & \textbf{18}\\
& Norm. AUC (Std $\downarrow$) & 0.345 & 0.343 & \textbf{0.215}\\
\midrule
\multirow{4}{*}{\textbf{TabArena}\cite{erickson_tabarena_2025}} 
& \#Wins & 1 & 6 & \textbf{13} \\
& AUC Change (\%)  & - & $0.166$ & $\mathbf{0.462}$ \\
& \#Leaps & - & 3 & \textbf{7}\\
& Norm. AUC (Std $\downarrow$) & 0.223 & 0.195 & \textbf{0.178}\\
\midrule
\multirow{4}{*}{\textbf{PMLB}\cite{olson2017pmlb}} 
& \#Wins & 15 & 8 & \textbf{37} \\
& AUC Change (\%)  & - & $0.016$ & $\mathbf{0.332}$ \\
& \#Leaps & - & 1 & \textbf{8}\\
& Norm. AUC (Std $\downarrow$) & 0.210 & 0.209 & \textbf{0.188}\\
\bottomrule
\end{tabular}
\end{adjustbox}
\label{table:tfm-only}
\end{table*}

\textbf{\model\ outperforms Fine-Tuning.} To further investigate the impact of finetuning versus our \model\ approach, in Table \ref{table:tfm-only}, we compare these two methods against the original TabPFN V2 model in isolation. Across the three benchmarks, we report the number of outright wins among the three TabPFN variants and the relative percentage change in mean raw AUC from the frozen model. Additionally, we report a metric we refer to as the \textit{number of leaps}, defined as the number of datasets for which the adapted TabPFN surpasses at least one conventional baseline that the original TabPFN-V2 did not outperform. This metric captures how often the method not only improves upon TabPFN-V2 incrementally, but improves it enough to surpass alternative model classes. This metric is key to our analysis, given that a major motivation for this work is the observation that, although TFMs are competitive with traditional methods, they do not consistently outperform them. In Table \ref{table:tfm-only}, we can see that applying \model\ not only significantly improves the average change in AUC over finetuning alone, but also dominates in the number of wins and, importantly, the number of leaps. In the most extreme case, TabPertNet, finetuning has almost no effect on performance, while applying \model\ leads to $18$ leaps, such that TabPFN would now outperform at least one alternative method on $25.7\%$ of datasets in the benchmark. Additionally, we report the standard deviation of normalized AUC scores for each model, which measures the dispersion of performance across datasets. We observe that the robust training process of \model\ reduces this cross-dataset variability.

\textbf{Generalization Across TFM Backbones.} To evaluate the generality of the proposed training framework, we apply \model\ to the open-source TabICL V2 classifier \citep{qu_tabiclv2_2026}, a TFM with a different architecture from TabPFN V2. We use the same domain-adaptation and robust training pipeline, without modifying the TabICL architecture, and compare against the frozen backbone and compute-matched fine-tuning on TabPertNet, TabArena, and PMLB. Table \ref{table:tabicl-main} summarizes normalized AUC, rank-1 wins, and mean rank across the three benchmarks. Both \model\ variants achieve the highest mean normalized AUC among the evaluated methods, and TabICL with \model\ improves over the original TabICL on all three benchmarks (Appendix \ref{app:tabicl}). The improvement over the frozen TabICL is significant under a one-sided Wilcoxon signed-rank test on per-dataset normalized AUC ($p=0.032$), whereas the corresponding improvement from fine-tuning is not ($p=0.148$). The direct comparison of TabICL with \model\ against fine-tuning favors \model\ in mean normalized AUC, but is not statistically significant (mean gap $0.018$; 95\% bootstrap confidence interval $[-0.005,0.043]$). These results demonstrate that the benefits of \model\ are not specific to TabPFN.

\begin{table*}[t]
\centering
\caption{Generalization of \model\ to the TabICL V2 backbone across TabPertNet, TabArena, and PMLB. Normalized AUC and ranks are computed over the expanded comparison of 11 models (including five classical baselines); selected models are shown. Scores are pooled across 177 datasets with defined normalized AUC and use a common normalization across all evaluated models.}
\setlength{\tabcolsep}{6pt}
\small
\begin{tabular}{lccc}
\toprule
\textbf{Model} & \textbf{Mean Norm. AUC} & \textbf{Rank-1 Wins} & \textbf{Mean Rank} \\
\midrule
TabICL V2 (\model) & \textbf{0.868} & 21 & 4.23 \\
TabPFN V2 (\model) & 0.860 & 11 & 4.73 \\
TabICL V2 (FT) & 0.850 & 15 & 4.37 \\
TabICL V2 & 0.837 & 18 & 4.67 \\
TabPFN V2 (FT) & 0.810 & 2 & 5.37 \\
TabPFN V2 & 0.806 & 1 & 5.45 \\
CatBoost & 0.743 & 8 & 6.21 \\
\bottomrule
\end{tabular}
\label{table:tabicl-main}
\end{table*}

\textbf{Isolating Domain Adaptation and Robustness.} We further ablate the two components of \model\ by training TabPFN V2 using only the domain-adapted reference distribution (DA-only), or only robust parameter selection (Robust-only). In Table \ref{table:components}, each component independently improves upon the frozen model, while the full approach achieves the highest mean AUC on five of the six benchmarks. Across 236 datasets, both DA-only and Robust-only improve normalized AUC significantly relative to the frozen TabPFN ($p=10^{-4}$ and $p=0.028$, respectively, using one-sided paired Wilcoxon tests). The effects are approximately additive on average, although the relative contribution varies by domain: domain adaptation accounts for most of the improvement on PMLB, while the two components complement one another on Engineering. Additional analysis is provided in Appendix \ref{app:components}.

\begin{table*}[t]
\centering
\caption{Ablation of domain adaptation (DA) and robust SCM parameter selection, reporting mean test AUC for TabPFN V2 across six benchmarks. The full method combines both components.}
\setlength{\tabcolsep}{7pt}
\small
\begin{tabular}{lcccc}
\toprule
\textbf{Benchmark} & \textbf{Base} & \textbf{DA-only} & \textbf{Robust-only} & \textbf{\model} \\
\midrule
PMLB & 0.9088 & \textbf{0.9121} & 0.9109 & 0.9119 \\
TabPertNet & 0.8019 & 0.8093 & 0.8113 & \textbf{0.8185} \\
TabArena & 0.8725 & 0.8738 & 0.8743 & \textbf{0.8765} \\
Genomics & 0.9675 & 0.9699 & 0.9679 & \textbf{0.9713} \\
Healthcare & 0.8915 & 0.8937 & 0.8938 & \textbf{0.8951} \\
Engineering & 0.9402 & 0.9403 & 0.9386 & \textbf{0.9422} \\
\midrule
\textbf{Pooled} & 0.8780 & 0.8819 & 0.8818 & \textbf{0.8851} \\
\bottomrule
\end{tabular}
\label{table:components}
\end{table*}

\textbf{Compute-Matched Fine-Tuning.} To separate the benefit of the training distribution from training duration, we additionally train TabPFN V2 on real reference datasets with the same batch size and number of training datasets and optimizer updates as the corresponding \model\ runs. Both approaches exhibit small gains at early checkpoints, but their trajectories diverge with additional training. Across TabArena, PMLB, and TabPertNet, fine-tuning eventually falls below the frozen backbone, while \model\ continues to improve. On TabArena, for example, the selected \model\ checkpoint from a training run with 30k generated-dataset exposures yields a mean raw AUC gain of $+0.0040$ relative to the frozen model, compared with $-0.0844$ for fine-tuning evaluated at the 30k checkpoint. The full checkpoint comparisons are reported in Appendix \ref{app:compute-matched}.

\textbf{Learned SCM Parameters and Generator Coverage.} The learned adversarial distributions vary substantially across benchmarks (Appendix \ref{app:scm-results}). For example, Genomics favors generators with many numerical features, whereas Healthcare favors small, predominantly categorical datasets with higher noise. The Engineering benchmark instead selects larger, dense generators with numerical features. These differences illustrate the importance of adapting both dataset-level properties and the underlying data-generating mechanisms. However, domain adaptation is necessarily limited by the expressiveness of the available SCM family. On a foreign-exchange benchmark with pronounced temporal structure, the original non-temporal generators provide poor coverage of the real-data distribution, and \model\ does not improve upon the frozen TabPFN. Extending the generator family with a temporal SCM improves alignment and yields a modest gain in AUC. This case illustrates both a limitation of the current generator space and the potential of domain-specific synthetic generation (Appendix \ref{app:forex}).

\section{Conclusion}
We introduce \model, a model-agnostic framework for robust, domain-adapted continued training of TFMs. We demonstrate its ability to improve TabPFN-V2 using less than $0.1\%$ of the original pretraining data, across six different tabular benchmarks. We further improve the architecturally distinct TabICL V2 classifier across three major benchmarks, demonstrating the generality of our approach. We show that constraining the robust training process to generate synthetic data that is similar to real-world data can be done efficiently, and that both domain adaptation and robust parameter selection independently contribute to performance. Compute-matched comparisons further distinguish the gains from training on robust, domain-adapted synthetic data from simply training for longer on real datasets. The foreign-exchange experiments further highlight the role of the synthetic generator family: extending its support to temporal data improves domain coverage and enables further adaptation. Our framework can also be extended to other TFMs and regression tasks. The \model\ algorithm is highly-parallelizable, enabling large-scale training and efficient parameter search. Together, these results demonstrate how robust optimization and flexible SCM-based synthetic data generation can improve TFM performance across diverse domains.
\begin{ack}
\textbf{Funding.} This work was supported by a research grant from the MIT--IBM Computing Research Lab.

\textbf{Competing interests.} The authors declare no competing interests.

\newpage

\end{ack}

\bibliographystyle{plain}
\bibliography{neurips_2026}

%%%%%%%%%%%%%%%%%%%%%%%%%%%%%%%%%%%%%%%%%%%%%%%%%%%%%%%%%%%%

\appendix

\section{Details for Domain-Adapted TFMs}\label{app:da}
In this Section, we provide a detailed overview of the domain-adaptation process introduced in Section \ref{sec:da}. We begin by defining the set of meta-features used to encode tabular datasets, then define our surrogate reward for the Bayesian Optimization process to maximize partial coverage of the real data distribution, and finally introduce our algorithm for constructing the global mixture over synthetic data generators. Although the overall pipeline involves multiple steps, the meta-features we define are inexpensive to compute, and we are able to solve the global mixture optimization efficiently via the Sinkhorn algorithm. As a result, we found that for all of our experiments, we can recover the approximate reference distribution over synthetic data generators in under 10 minutes, using a single A100 GPU on standard hardware. Further, this process only needs to be run once for each domain, since the same reference distribution can be re-used across many training runs of the \model\ algorithm. We summarize the pseudocode for our domain-adaptation pipline in Algorithms \ref{alg:da} and \ref{alg:domain-param-search}.

\textbf{Meta-Feature Extraction.} For a tabular dataset $(X,y)$ with $n$ samples and $p$ input features, we construct a meta-feature embedding
\[
\phi(X,y)=
\begin{bmatrix}
\phi_{\mathrm{schema}}(X)\\
\phi_{\mathrm{geom}}(X)\\
\phi_{\mathrm{learn}}(X,y)
\end{bmatrix}
\in \mathbb{R}^d.
\]
The representation is designed to separate three sources of variation: coarse dataset schema and marginals, affine-invariant geometric structure of the numeric covariates, and inexpensive proxies for supervised learnability. In the classification setting used in our experiments, the corresponding block dimensions are $(d_{\mathrm{schema}},d_{\mathrm{geom}},d_{\mathrm{learn}})=(24,24,16)$.

Throughout, for any finite vector $v=(v_1,\dots,v_m)$, we write
\[
S(v)=\bigl(\mathrm{mean}(v),\ \mathrm{std}(v),\ Q_{0.25}(v),\ \mathrm{median}(v),\ Q_{0.75}(v)\bigr)
\]
for the five-dimensional summary operator applied repeatedly below. The schema block contains the global descriptors:
$\log(1+p)$, categorical feature ratio, total missing fraction,
together with the fraction of categorical features that are binary, and the summaries $S(\text{per-feature missing rate})$, $S(\text{numeric unique ratio})$, $S(\text{categorical unique ratio})$, $S(\text{categorical normalized entropy})$. Note that for this work, we do not encode schema information regarding the number of samples in the dataset, since we currently fix the number of samples generated for synthetic datasets to be a constant during model training. In future work, we plan to include these features of the dataset schema, but we exclude them for now for training stability purposes. These quantities preserve coarse structural properties of the dataset that should not be removed by downstream normalization.

Let $X_{\mathrm{num}}\in\mathbb{R}^{n\times p_{\mathrm{num}}}$ denote the numeric submatrix. For each numeric feature $j$, we compute the mean $\mu_j$ and standard deviation $\sigma_j$ using observed entries only, impute missing values by $\mu_j$, and form the standardized column
\[
\widetilde X_{ij}=\frac{X_{ij}-\mu_j}{\sigma_j}.
\]
All geometric features are computed from $\widetilde X_{\mathrm{num}}$, making this block invariant to per-feature affine transformations $X_{ij}\mapsto a_j X_{ij}+b_j$ with $a_j>0$. The geometric block contains $S(|\text{feature skewness}|)$, $S(\text{feature kurtosis})$, and
$S(\text{feature outlier rate})$, where the outlier rate is the fraction of observed standardized entries satisfying $|\widetilde X_{ij}|>3$. To capture dependence structure, we form the correlation matrix $R$ of the standardized numeric features and include
\[
S\!\bigl(|R_{jk}|:j<k\bigr),\qquad \max_{j<k}|R_{jk}|,
\]
together with three spectral summaries,
\[
\frac{\lambda_1}{\sum_\ell \lambda_\ell},\qquad
\frac{\sum_{\ell=1}^{\min(5,r)}\lambda_\ell}{\sum_\ell \lambda_\ell},\qquad
\exp\!\Bigl(-\sum_\ell \pi_\ell\log \pi_\ell\Bigr),
\]
where $\lambda_1\geq \lambda_2\geq \cdots\geq 0$ are the eigenvalues of $R$, $r$ is the rank, and $\pi_\ell=\lambda_\ell/\sum_k\lambda_k$. The final term is the entropy-based effective rank.

The learnability block augments these unsupervised statistics with inexpensive surrogates for task difficulty. Specifically, it includes \#classes, $H(Y)$, $ \max_c \Pr(Y=c)$,
$\min_c \Pr(Y=c)$, 
where $H(Y)=-\sum_c \Pr(Y=c)\log \Pr(Y=c)$. It further includes the summaries $S(\eta_j^2)$ over numeric features, and $S(V_j)$ over categorical features, 
where $\eta_j^2$ is the ANOVA effect size between numeric feature $j$ and the class label, and $V_j$ is Cram\'er's $V$ between categorical feature $j$ and the class label. Finally, we include two global proxies for separability: leave-one-out nearest-centroid accuracy, and the ratio
\[
\frac{\text{mean pairwise distance between class centroids}}{\text{mean within-class distance}}.
\]

Given a collection of reference datasets $\{(X^{(i)},y^{(i)})\}_{i=1}^N$, we first compute raw embeddings
\[
u^{(i)}=\phi(X^{(i)},y^{(i)})\in\mathbb{R}^d.
\]
For each coordinate $k$, we then estimate the reference-set mean and variance,
\[
\mu_k=\frac{1}{N}\sum_{i=1}^N u_k^{(i)},\qquad
s_k^2=\frac{1}{N}\sum_{i=1}^N (u_k^{(i)}-\mu_k)^2,
\]
and standardize each embedding coordinate according to
\[
z_k^{(i)}=\frac{u_k^{(i)}-\mu_k}{s_k}.
\]
Any generated dataset is mapped into the same latent space using the \emph{same} $(\mu_k,s_k)$ estimated from the reference collection. After this per-coordinate standardization, we apply an explicit block reweighting. Writing
\[
z=
\begin{bmatrix}
z_{\mathrm{schema}}\\
z_{\mathrm{geom}}\\
z_{\mathrm{learn}}
\end{bmatrix},
\qquad
z_b\in\mathbb{R}^{d_b},
\]
we define the final transformed representation by
\[
\widehat z_b=\sqrt{\frac{w_b}{d_b}}\,z_b,
\]
where $w_b>0$ is a user-specified block weight. Distances are then evaluated in the transformed space:
\[
\|\widehat z-\widehat z'\|_2^2
=
\sum_b \frac{w_b}{d_b}\,\|z_b-z_b'\|_2^2.
\]
This normalization prevents large blocks from dominating the geometry solely by virtue of dimensionality. If each standardized coordinate has variance approximately one and within-block correlations are not too strong, then
\[
\mathbb{E}\bigl[\|z_b\|_2^2\bigr]\approx d_b,
\qquad
\mathbb{E}\bigl[\|\widehat z_b\|_2^2\bigr]
=
\frac{w_b}{d_b}\,\mathbb{E}\bigl[\|z_b\|_2^2\bigr]
\approx w_b.
\]
Thus the expected squared contribution of block $b$ is controlled by $w_b$ rather than by its raw dimension. In practice, we set $w_b = \frac{1}{3}$ for each of the three categories, balancing the contribution of each to the distance metric.

\textbf{Parameter Search for Partial Coverage.} With our dataset encoder and distance metric defined, we now introduce our search method for finding parameterizations of synthetic data generators that align with real-world data from some reference corpus. Since this is an unconstrained black-box optimization problem over a discrete search space, we use Bayesian Optimization to find parameters that cover parts of the reference corpus well \cite{frazier2018tutorial}. To accomplish this, we now define our surrogate objective function that rewards partial coverage of the embedding space spanned by the reference corpus. For a given reference corpus of datasets, let $\{\mathbf{x}_i\}_{i=1}^n \subset \mathbb{R}^d$ denote its standardized reference embeddings after the same preprocessing and block reweighting described in the previous section. A trial parameter vector $\theta$ induces a \emph{set} of synthetic datasets; in the implementation, we generate $m$ datasets from $\phi \sim p(\Phi;\theta)$, embed them, and obtain a cloud $\{\mathbf{x}_j(\theta)\}_{j=1}^m \subset \mathbb{R}^d$. The search objective is then defined in terms of how well this cloud covers the reference embeddings.

For each reference point $\mathbf{x}_i$, the we compute a soft nearest-neighbor distance to the generated cloud,
\begin{equation}\label{eq:param-dist-metric}
d_i(\theta)
=
-\tau \log\!\left(
\frac{1}{m}\sum_{j=1}^m
\exp\!\left(-\frac{\|\mathbf{x}_i-\mathbf{x}_j(\theta)\|_2}{\tau}\right)
\right),
\end{equation}
with temperature $\tau>0$. Distances are Euclidean and $\tau=1$ by default. As $\tau \to 0$, $d_i(\theta)$ approaches $\min_j \|\mathbf{x}_i-\mathbf{x}_j(\theta)\|_2$, while for larger $\tau$ it becomes a smoother average over nearby generated embeddings. This gives a differentiable surrogate for pointwise coverage without requiring an exact minimum.

The scalar objective returned to the optimizer at iteration $t$ is a residual-weighted average of these pointwise distances,
\begin{equation}
f_t(\theta)
=
-\sum_{i=1}^n r_i^{(t)} d_i(\theta),
\qquad
r_i^{(t)} \ge 0,
\qquad
\sum_{i=1}^n r_i^{(t)} = 1.
\end{equation}
Maximizing $f_t$ is therefore equivalent to minimizing a weighted coverage loss. The weights $r_i^{(t)}$, which we refer to as the \emph{residuals}, capture which parts of the reference data remain poorly covered by the synthetic data generators found so far. To define these residual weights, we track the coverage distance $c_i^{(t)}$ for each reference point. After evaluating a new parameter vector $\theta_t$, the candidate distances $d_i(\theta_t)$ are merged with the existing coverage state by a hard minimum,
\begin{equation}
c_i^{(t+1)}
=
\begin{cases}
d_i(\theta_t), & t=0,\\
\min\bigl (c_i^{(t)},\, d_i(\theta_t)\bigr), & t \ge 1.
\end{cases}
\end{equation}
Thus $c_i^{(t)}$ records the best coverage attained so far by any previously evaluated generator in that cluster. The residual weights for the next trial are then recomputed from the updated coverage distances. The update is the linear normalization
\begin{equation}
r_i^{(t+1)}
=
\frac{c_i^{(t+1)}}{\sum_{\ell=1}^n c_\ell^{(t+1)}}.
\end{equation}
Reference points that are already well covered have small $c_i^{(t+1)}$ and therefore receive little residual mass, whereas persistently under-covered points retain large weights. The initial residual state is uniform, $r_i^{(0)} = 1/n$. Consequently, early trials try to improve average coverage of the whole space, while later trials are pushed toward the portions of embedding space that remain inadequately represented. This is why we refer to the score as a \emph{partial coverage} surrogate, since a single generator is not required to match the entire reference corpus by itself. Instead, the sequential parameter search encourages successive generators to cover complementary regions of the target embedding distribution. In practice, as we can see in Figure \ref{fig:three-subfigures}, there are no generators the adequately cover the entire space spanned by the reference corpus. Instead, by finding generators that partially cover this space and combining them, we are able to sufficiently cover the reference corpus.

\textbf{Solving the Global Mixture Problem.} After the parameter search from the previous section is performed, we obtain $T$ candidate generators. Generator $t$ is represented by a cloud of embedded synthetic datasets $\{\boldsymbol{x}_j(\theta_t)\}_{j=1}^{m_t} \subset \mathbb{R}^d$, while the target collection of real datasets is represented by reference embeddings $\{\boldsymbol{x}_i\}_{i=1}^n \subset \mathbb{R}^d$. In practice, we set $m_t=m$ as a constant for consistency across trials. We want to fit a global distribution over generators by choosing mixture weights $\boldsymbol{\alpha}=(\alpha_1,\dots,\alpha_T)$ on the simplex that minimizes the transport distance between this global distribution over synthetic data generators and the distribution over the reference datasets.

The reference embeddings define the empirical measure
\begin{equation}
\mu
=
\frac{1}{n}\sum_{i=1}^n \delta(\boldsymbol{x}_i),
\end{equation}
and a candidate generator mixture induces
\begin{equation}
\nu_{\boldsymbol{\alpha}}
=
\sum_{k=1}^T \alpha_t
\left(
\frac{1}{m_t}\sum_{j=1}^{m_t} \delta(\boldsymbol{x}_j(\theta_t))
\right).
\end{equation}
Thus each synthetic table embedding produced by generator $k$ receives mass $\alpha_t/m_t$. We compare $\mu$ and $\nu_{\boldsymbol{\alpha}}$ using squared Euclidean cost $c(\boldsymbol{x},\boldsymbol{y})=\|\boldsymbol{x}-\boldsymbol{y}\|_2^2$ and the entropically regularized optimal transport objective
\begin{equation}
\mathrm{OT}_{\varepsilon}(\mu,\nu_{\boldsymbol{\alpha}})
=
\min_{\boldsymbol{\Pi}\in U(\boldsymbol{a},\boldsymbol{b}(\boldsymbol{\alpha}))}
\langle \boldsymbol{\Pi}, \boldsymbol{C}\rangle
+ \varepsilon \sum_{i,j} \Pi_{ij}(\log \Pi_{ij}-1),
\end{equation}
where $\varepsilon>0$ is the regularization parameter, $\boldsymbol{a}_i=1/n$, $\boldsymbol{b}(\boldsymbol{\alpha})$ collects the per-sample masses $\alpha_t/m_t$, and $U(\boldsymbol{a},\boldsymbol{b})$ is the set of couplings with marginals $\boldsymbol{a}$ and $\boldsymbol{b}$. The entropic term makes the problem strictly convex in the transport plan and enables efficient solution by Sinkhorn matrix scaling \cite{cuturi2013sinkhorn}.

Rather than solving the transport problem from scratch by general-purpose linear programming, we use the Sinkhorn algorithm in the log domain. Writing $C_{ij}=\|\boldsymbol{x}_i-\boldsymbol{y}_j\|_2^2$, the dual potentials are updated as
\begin{align}
f_i &\leftarrow \varepsilon\!\left(\log a_i - \log \sum_j \exp\!\left(\frac{-C_{ij}+g_j}{\varepsilon}\right)\right),\\
g_j &\leftarrow \varepsilon\!\left(\log b_j - \log \sum_i \exp\!\left(\frac{-C_{ij}+f_i}{\varepsilon}\right)\right),
\end{align}
which yields the transport plan $\Pi_{ij}=\exp(( -C_{ij}+f_i+g_j)/\varepsilon)$. This log-domain form is numerically stable and gives an efficient inner solver for repeated OT evaluations.

We then solve the outer mixture problem by minimizing the regularized transport cost with respect to $\boldsymbol{\alpha}$:
\begin{equation}
\min_{\boldsymbol{\alpha}\in\Delta^{T-1}}
\mathrm{OT}_{\varepsilon}(\mu,\nu_{\boldsymbol{\alpha}}).
\end{equation}
In practice, we reparameterize $\boldsymbol{\alpha}$ by unconstrained logits $\boldsymbol{\eta}\in\mathbb{R}^T$ with $\alpha_t=\exp(\eta_t)/\sum_{\ell=1}^T \exp(\eta_\ell)$, and optimize $\boldsymbol{\eta}$ by gradient descent. Each outer step constructs the sample-weight vector $\boldsymbol{b}(\boldsymbol{\alpha})$, calls the Sinkhorn solver to evaluate $\mathrm{OT}_{\varepsilon}(\mu,\nu_{\boldsymbol{\alpha}})$, and backpropagates through that solution to update the logits. In this way, Sinkhorn serves as the inner OT subroutine inside the global mixture optimization. The result is a set of mixture weights over generators whose aggregated embedding distribution best matches the reference embeddings under the entropically regularized transport geometry.

\begin{algorithm}[t]
\caption{Pseudocode for Domain-Adapted TFMs}\label{alg:da}
\begin{algorithmic}[1]
\Require Reference datasets $\mathcal{D}_{\mathrm{ref}}=\{(\boldsymbol{X}^{(i)},\boldsymbol{y}^{(i)})\}_{i=1}^N$, parameter space $\Theta$, trials $T$, datasets per trial $m$, soft-coverage temperature $\tau$, transport regularization $\varepsilon$, Sinkhorn iterations $L$, outer iterations $O$
\State Compute reference embeddings $\{\boldsymbol{x}_i\}_{i=1}^N$ from $\mathcal{D}_{\mathrm{ref}}$
\State Standardize and reweight the embedding coordinates using statistics estimated from $\mathcal{D}_{\mathrm{ref}}$
\State Initialize an empty list of candidate generators and generated embedding clouds
\State Initialize residual weights uniformly as $\boldsymbol{r}^{(0)} = \frac{1}{N}$
\State $\{(\theta_t, \{\mathbf{x}_j(\theta_t)\}_{j=1}^m)\}_{t=1}^T \gets \mathrm{DomainParamSearch}(\{\boldsymbol{x}_i\}_{i=1}^N,\mathbf{r},\Theta,m,\tau,T)$
\State Form the empirical reference measure $\mu$ from $\{\boldsymbol{x}_i\}_{i=1}^N$
\State For each candidate generator $\theta_t$, form the empirical measure supported on its generated cloud $\mathcal{Y}_t = \frac{1}{m}\sum_{j=1}^{m} \delta(\boldsymbol{x}_j(\theta_t))$
\State Define $\nu_{\boldsymbol{\alpha}} = \sum_{t=1}^T \alpha_t \mathcal{Y}_t$ as the global mixture over synthetic data generators
\State Solve $\boldsymbol{\alpha}^\star \in \arg\min_{\boldsymbol{\alpha}\in\Delta^{M-1}} \mathrm{OT}_{\varepsilon}\!\bigl(\mu,\nu_{\boldsymbol{\alpha}}\bigr)$ using gradient-based optimization over mixture logits, with Sinkhorn$(\mu, \nu_{\boldsymbol{\alpha}}, \varepsilon,L)$ as the inner OT solver, for $O$ outer iterations
\State \textbf{return} Global mixture over generator parameters $\boldsymbol{\alpha}^\star$
\end{algorithmic}
\end{algorithm}

\begin{algorithm}[t]
\caption{\textsc{DomainParamSearch}}\label{alg:domain-param-search}
\begin{algorithmic}[1]
\Require Reference embeddings $\{\boldsymbol{x}_i\}_{i=1}^N$, initial residuals $\boldsymbol{r}^{(0)}$, parameter space $\Theta$, datasets per trial $m$, soft-coverage temperature $\tau$, trials $T$
\State Initialize cover distances $\boldsymbol{c}^{(0)} = (+\infty,\dots,+\infty)$
\State Initialize history $\mathcal{H}^{(0)} = \emptyset$
\For{$t=1,\dots,T$}
    \State $\theta_t \gets \mathrm{BayesOPT}(\mathcal{H}^{(t-1)}, \Theta)$
    \State Generate $m$ synthetic datasets from $\theta_t$, compute their embeddings, and collect them as $\{\mathbf{x}_j(\theta_t)\}_{j=1}^m$
    \State Compute the pointwise soft coverage distances $\{d_i(\theta_t)\}_{i=1}^N$ as in Equation \ref{eq:param-dist-metric}
    \State Compute the partial coverage reward $f_t(\theta_t) = -\sum_{i=1}^N r_i^{(t-1)} d_i(\theta_t)$
    \For{$i=1,\dots,N$}
        \State $c_i^{(t)} \gets \min\{c_i^{(t-1)}, d_i(\theta_t)\}$
    \EndFor
    \For{$i=1,\dots,N$}
        \State $r_i^{(t)} \gets c_i^{(t)} \big/ \sum_{\ell=1}^N c_\ell^{(t)}$
    \EndFor
    \State Update the BO history as $\mathcal{H}^{(t)} = \mathcal{H}^{(t-1)} \cup \{(\theta_t, f_t(\theta_t))\}$
\EndFor
\State \textbf{return} Parameters and generated dataset point clouds $\{(\theta_t,\{\mathbf{x}_j(\theta_t)\}_{j=1}^m)\}_{t=1}^T$
\end{algorithmic}
\end{algorithm}

\section{\model\ Algorithm Details}\label{app:rtfm}
We summarize \modelfull\ in Algorithms \ref{alg:rtfm} and \ref{alg:parameter-search}. A description of the algorithm is given in Section \ref{sec:rtfm}. To begin, we perform the domain-adaption search from given by Algorithm \ref{alg:da} to construct a reference distribution $P$ over SCM parameters that aligns the synthetic data with a given reference corpus of real data. We then alternate between a maximization stage, in which we search over SCM parameters to find those with large optimality gaps, as in Equation \ref{eq:dro-objective}, and a minimization stage, where we sample generators from an adversarial distribution to continue training the TFM. In Algorithm \ref{alg:parameter-search}, we reference an abstract function SuggestParameters which represents any black-box optimization algorithm. In this case, we use the Optuna python package \cite{akiba_optuna_2019} and the Tree-structed Parzen Estimator \cite{watanabe_tree-structured_2023} algorithm, which belongs to the class of Bayesian Optimization algorithms \cite{frazier2018tutorial}. At each iteration, this algorithm uses the information gained from the previous trials (the estimated optimality gaps) to propose a new parameterization, balancing exploration and exploitation to efficiently find parameters corresponding to large optimality gaps. To begin the minimization stage, we first construct the distribution $Q \in \Delta_{\mathcal{P}}$. Rather than constructing $\rho$ as a parameter itself, we introduce a scalar parameter $c \in (0,1)$ and define $\rho = c \cdot \rho_{\text{max}}$, such that $\rho$ is a given as a fraction of the maximum possible KL-divergence of $Q$ given the reference distribution $P$. With $\rho_{\text{max}}$ specified, we can find $\eta$ such that $D_{KL}(Q(\eta)||P) = \rho_{\text{max}}$ using one-dimensional root finding  (e.g. bisection). Then we generate each dataset in a training batch by sampling a parameter $\theta_i \sim Q$ and generator $\phi \sim p(\Phi; \theta_i)$. We train the model for a fixed number of iterations and then return to the maximization stage. In our implementation, after five epochs, we incorporate the original TabPFN model as another baseline model. This can potentially address regions of the parameter space where the model has reduced its performance by focusing on improving in other regions.

The \model\ algorithm is highly parallelizable, both in the dataset generation steps and the parameter search. For a fixed parameterization $\theta$, each dataset generator $\phi \sim p(\Phi; \theta)$ is sampled independently from the same distribution, such that all datasets generated for a given batch can be created concurrently. The ParameterSearch algorithm can be completely parallelized within each trial. In our implementation, for each trial, we generate a batch of datasets and then fit and evaluate each model and dataset pair concurrently. For example, in our experiments, we set $n_d=20$ and used $e=7$ baseline models, resulting in $n_{ds} * e = 140$ separate processes distributed across all available cores. In practice, the domain-adaptation stage is lightweight, while the overall training cost scales approximately linearly with the number of max-min iterations; we provide benchmark-specific compute details in Appendix \ref{app:experiments}.
\begin{algorithm}[t]
\caption{\modelfull\ (\model)}\label{alg:rtfm}
\begin{algorithmic}[1]
\Require Initial TFM weights $\mathbf{W}$, baseline estimators $\{f_1, \dots, f_e\}$, $n_e$, $n_{iter}$, $n_t$, $n_d$, $c \in (0,1)$, reference distribution $P$ from Algorithm \ref{alg:da}, parameter space $\Theta$
\State $\{(\theta_i, \delta_i)\}_{i=1}^{n_{trials}} \gets \text{ParameterSearch}(\Theta, \mathbf{W}, \{f_1, \dots, f_e\}, n_t, n_d)$
\State $\rho \gets c \cdot \rho_{\text{max}}$
\State $\eta \gets \text{RootFinding}(\rho, P, \mathbf{\delta})$ \Comment{Find correct temperature for distribution}
\State Construct distribution $Q_i \propto p_i \exp(\eta \cdot \delta_i)$
\For{$e \in [n_e]$}
\For{$t \in [n_{iter}]$} 
% split into train/test sets
\State $\theta_1,\dots,\theta_{n_b} \sim Q$ \Comment{Sample parameters}
\State $\phi_1,\dots, \phi_{n_b} \sim p(\Phi ; \theta_1), \dots,p(\Phi;\theta_{n_b})$
\State $\{(x^i, y^i)\}_{i=1}^{n_b} \sim p(\phi_1),\dots, p(\phi_{n_b})$ \Comment{Sample synthetic datasets}
\State $\{x^i_{train}, x^i_{test}, y^i_{train}, y^i_{test}\}_{i=1}^{n_b} \gets \text{TrainTestSplit}(\{(x^i, y^i)\}_{i=1}^{n_b})$
\State $l(g_{\mathbf{W}}) \gets \mathcal{L}_{PFN}(g_{\mathbf{W}}, \{x^i_{train}, x^i_{test}, y^i_{train}, y^i_{test}\}_{i=1}^{n_b})$
\State $\mathbf{W} \gets \text{update}(\mathbf{W}, l(g_{\mathbf{W}}))$ \Comment{Gradient step}
\EndFor
\State $\{(\theta_i, \delta_i)\}_{i=1}^{n_{trials}} \gets \text{ParameterSearch}(\mathbf{W}, \{f_1, \dots, f_e\}, n_t, n_d)$
\State $\eta \gets \text{RootFinding}(H_{min}, \delta)$
\State Construct distribution $Q_i \propto p_i \exp(\eta \cdot \delta_i)$
\EndFor
\State \textbf{Return} Robust TFM model $g_{\mathbf{W}}$
\end{algorithmic}
\end{algorithm}

\begin{algorithm}[h]
\caption{ParameterSearch}\label{alg:parameter-search}
\begin{algorithmic}[1]
\Require TFM weights $\mathbf{W}$, baseline estimators $\{f_1, \dots, f_e\}$, num. trials $n_t$, num. datasets $n_d$
\State $D_{\theta} = \{\}$
\For{$i \in [n_t]$}
\State $\theta_i \gets \text{SuggestParameters}(D_{\theta})$ \Comment{Black-box optimizer proposes new parameter}
\For{$j \in [n_d]$}
\State $\phi_j \sim p(\Phi ; \theta_i)$
\State $x^j_{train}, x^j_{test}, y^j_{train}, y^j_{test} \sim p(\phi_j)$
\State $l_j(g_{\mathbf{W}}) \gets \mathcal{L}_{PFN}(g_{\mathbf{W}}, x^j_{train}, y^j_{train}, x^j_{test}, y^j_{test})$
\State $l_b \gets \infty$
\For{$f_k \in \{f_1, \dots, f_e\}$}
% Emphasize here that some (most) of these models need to be fit
\State $l_j(f_k) \gets \mathcal{L}_{PFN}(f_k, x^j_{train}, y^j_{train}, x^j_{test}, y^j_{test})$
\State $l_b \gets \min(l_b, l_j(f_k))$
\EndFor
\State $\delta_i^j \gets l_j(g_{\mathbf{W}}) - l_b$ \Comment{Estimate optimality gap}
\EndFor
\State $\delta_i \gets \frac{1}{n_d}\sum_{j=1}^{n_d}\delta_i^j$
\State $D_{\theta} \gets D_{\theta} \cup \{(\theta_i, \delta_i)\}$
\EndFor
\State \textbf{Return} Parameters and estimated optimality gaps $D_{\theta}$
\end{algorithmic}
\end{algorithm}

\section{Theoretical Properties}\label{app:proofs}
In this section, we provide the proof that the optimal distribution for the maximizer in equation \ref{eq:dro-objective} is a softmax distribution such that $q_i^* \propto p_i\exp(\eta \cdot \delta_{\theta_i}(\mathbf{W}))$. Further, we show that in the absence of a reference distribution, if we take $P$ to be uniform, our formulation can be written as an entropy-constrained DRO, where $Q$ is constrained to have a certain minimum entropy.

\begin{proposition}
Let $\delta=(\delta_{\theta_1}(\mathbf W),\dots,\delta_{\theta_n}(\mathbf W))\in\mathbb R^n$ and let
$P=(p_1,\dots,p_n)\in\Delta_n$ satisfy $p_i>0$ for all $i$. Consider the distributionally robust inner problem
\[
\max_{Q\in\Delta_n}\ \sum_{i=1}^n q_i\delta_{\theta_i}(\mathbf W)
\quad\text{s.t.}\quad
D_{\mathrm{KL}}(Q\|P)\le \rho,
\]
Let $\mathcal A=\argmax_{i\in[n]} \delta_{\theta_i}(\mathbf W)$, and
$P(\mathcal A)=\sum_{i\in\mathcal A}p_i$. Assume that $\delta_{\theta_i}(\mathbf W)$ is not constant across $i$ and $0<\rho<-\log P(\mathcal A)$. Then the following properties hold.

\begin{enumerate}
    \item[(i)] The optimizer $Q^*$ is strictly positive, and there exists a unique temperature parameter $\eta>0$ such that
    \[
    q_i^*
    =
    \frac{
    p_i\exp\left(\delta_{\theta_i}(\mathbf W)/\eta\right)
    }{
    \sum_{j=1}^n p_j\exp\left(\delta_{\theta_j}(\mathbf W)/\eta\right)
    },
    \qquad i=1,\dots,n,
    \]
    where $\eta$ is determined implicitly by $D_{\mathrm{KL}}(Q^*\|P)=\rho$.

    \item[(ii)] If $P$ is the uniform distribution on $[n]$, then the KL constraint $D_{\mathrm{KL}}(Q\|P)\le \rho$ is equivalent to the entropy constraint $H(Q)\ge \log n-\rho$. In this case, the optimizer in part (i) reduces to the softmax distribution
    \[
    q_i^*
    =
    \frac{
    \exp\left(\delta_{\theta_i}(\mathbf W)/\eta\right)
    }{
    \sum_{j=1}^n \exp\left(\delta_{\theta_j}(\mathbf W)/\eta\right)
    },
    \qquad i=1,\dots,n,
    \]
    with $\eta>0$ chosen so that $H(Q^*)=\log n-\rho$.
\end{enumerate}
\end{proposition}

\begin{proof}
\textbf{Part (i)}. For notational simplicity, write $\delta_i=\delta_{\theta_i}(\mathbf W)$. The problem can be rewritten as
\[
\min_{Q\in\Delta_n}\ -\sum_{i=1}^n q_i\delta_i
\quad\text{s.t.}\quad
D_{\mathrm{KL}}(Q\|P)\le \rho.
\]
Since $P\in\Delta_n$ satisfies $p_i>0$ for all $i$ and
\[
D_{\mathrm{KL}}(P\|P)=0<\rho,
\]
Slater's condition holds. Therefore, the KKT conditions are necessary and sufficient.

The Lagrangian is
\[
\mathcal L(Q,\eta,\mu,\gamma)
=
-\sum_{i=1}^n q_i\delta_i
+
\eta\left(
\sum_{i=1}^n q_i\log\frac{q_i}{p_i}
-
\rho
\right)
+
\mu\left(\sum_{i=1}^n q_i-1\right)
-
\sum_{i=1}^n \gamma_i q_i,
\]
where $\eta\ge 0$, $\mu\in\mathbb R$, and $\gamma_i\ge 0$.

We first show that the KL constraint must be active. If the KL constraint were inactive, then $\eta=0$. In that case, the solution would have to maximize the linear objective over the simplex without an active KL constraint, and hence would place all mass on the set
\[
\mathcal A=\argmax_{i\in[n]}\delta_i.
\]
Among all distributions supported on $\mathcal A$, the one with smallest KL divergence from $P$ is the conditional distribution
\[
q_i=\frac{p_i}{P(\mathcal A)}\mathbf 1\{i\in\mathcal A\},
\]
whose KL divergence is
\[
D_{\mathrm{KL}}(Q\|P)
=
\sum_{i\in\mathcal A}\frac{p_i}{P(\mathcal A)}
\log\frac{p_i/P(\mathcal A)}{p_i}
=
-\log P(\mathcal A).
\]
Thus, if $\rho<-\log P(\mathcal A)$, no distribution supported entirely on $\mathcal A$ is feasible. Therefore the unconstrained maximizer is infeasible, so the KL constraint must be active and $\eta>0$.

For an interior optimum, complementary slackness gives $\gamma_i=0$. The stationarity condition is then
\[
-\delta_i
+
\eta\left(1+\log\frac{q_i}{p_i}\right)
+
\mu
=
0.
\]
Rearranging,
\[
\log\frac{q_i}{p_i}
=
\frac{\delta_i-\mu}{\eta}-1.
\]
Exponentiating gives
\[
q_i
=
p_i\exp\left(\frac{\delta_i}{\eta}\right)
\exp\left(-\frac{\mu}{\eta}-1\right).
\]
Enforcing $\sum_i q_i=1$ yields
\[
q_i(\eta)
=
\frac{
p_i\exp(\delta_i/\eta)
}{
Z(\eta)
},
\qquad
Z(\eta)
=
\sum_{j=1}^n p_j\exp(\delta_j/\eta).
\]
Since $p_i>0$ and $\eta>0$, this distribution is strictly positive.

It remains to show that there is a unique $\eta>0$ satisfying $D_{\mathrm{KL}}(Q(\eta)\|P)=\rho$. Let $E(\eta)=\sum_{i=1}^n q_i(\eta)\delta_i$. Then, For the distribution $Q(\eta)$,
\[
\log\frac{q_i(\eta)}{p_i}
=
\frac{\delta_i}{\eta}-\log Z(\eta),
\]
so
\[
D_{\mathrm{KL}}(Q(\eta)\|P)
=
\sum_{i=1}^n q_i(\eta)
\left(
\frac{\delta_i}{\eta}-\log Z(\eta)
\right)
=
\frac{E(\eta)}{\eta}-\log Z(\eta).
\]
Next,
\[
\frac{d}{d\eta}\log Z(\eta)
=
\frac{1}{Z(\eta)}
\sum_{i=1}^n p_i\exp(\delta_i/\eta)
\left(-\frac{\delta_i}{\eta^2}\right)
=
-\frac{E(\eta)}{\eta^2}.
\]
And,
\[
\frac{dE}{d\eta}
=
-\frac{\operatorname{Var}_{Q(\eta)}(\delta)}{\eta^2}.
\]
Therefore,
\begin{align*}
\frac{d}{d\eta}D_{\mathrm{KL}}(Q(\eta)\|P)
&=
\frac{1}{\eta}\frac{dE}{d\eta}
-
\frac{E(\eta)}{\eta^2}
-
\frac{d}{d\eta}\log Z(\eta) \\
&=
\frac{1}{\eta}\frac{dE}{d\eta}
-
\frac{E(\eta)}{\eta^2}
+
\frac{E(\eta)}{\eta^2} \\
&=
-\frac{\operatorname{Var}_{Q(\eta)}(\delta)}{\eta^3}.
\end{align*}
Since $\delta_i$ is not constant across $i$ and $Q(\eta)$ is strictly positive, we have
\[
\operatorname{Var}_{Q(\eta)}(\delta)>0
\]
for every $\eta>0$. We can then conclude that $\eta\mapsto D_{\mathrm{KL}}(Q(\eta)\|P)$ is strictly decreasing on $(0,\infty)$.

Further, we have $\lim_{\eta\to\infty}Q(\eta)=P$, so $\lim_{\eta\to\infty}D_{\mathrm{KL}}(Q(\eta)\|P)=0$. As $\eta\downarrow 0$, $Q(\eta)$ converges to $P$ conditioned on the set of maximizers $\mathcal A$, namely
\[
q_i(0)
=
\frac{p_i}{P(\mathcal A)}\mathbf 1\{i\in\mathcal A\},
\]
and therefore
\[
\lim_{\eta\downarrow 0}D_{\mathrm{KL}}(Q(\eta)\|P)
=
-\log P(\mathcal A).
\]
Thus, for every $0<\rho<-\log P(\mathcal A)$ there exists a unique $\eta>0$ such that
$D_{\mathrm{KL}}(Q(\eta)\|P)=\rho$. By the KKT conditions, this $Q(\eta)$ is the global optimizer.

\textbf{Part (ii)}. Suppose $P$ is uniform, so $p_i=1/n$ for all $i$. Then
\begin{align*}
D_{\mathrm{KL}}(Q\|P)
&=
\sum_{i=1}^n q_i\log\frac{q_i}{1/n} \\
&=
\sum_{i=1}^n q_i\log q_i
+
\sum_{i=1}^n q_i\log n \\
&=
\sum_{i=1}^n q_i\log q_i
+
\log n \\
&=
\log n - H(Q).
\end{align*}
Therefore, $D_{\mathrm{KL}}(Q\|P)\le \rho$ is equivalent to $\log n-H(Q)\le \rho$, or
$H(Q)\ge \log n-\rho$. Finally, substituting $p_i=1/n$ into the solution from part (i) gives
\[
q_i^*
=
\frac{
(1/n)\exp(\delta_i/\eta)
}{
\sum_{j=1}^n (1/n)\exp(\delta_j/\eta)
}
=
\frac{
\exp(\delta_i/\eta)
}{
\sum_{j=1}^n \exp(\delta_j/\eta)
},
\]
with $\eta>0$ chosen so that $D_{\mathrm{KL}}(Q^*\|P)=\rho$. Using $D_{\mathrm{KL}}(Q^*\|P)=\log n-H(Q^*)$, this is equivalently $H(Q^*)=\log n-\rho$. This proves the claim.
\end{proof}

\section{SCM Parameters}\label{app:scm}
We summarize the MLP-based SCM parameter space used in our experiments in Table \ref{table:params}. In this work, we instantiate SCMs as randomized MLPs, so the search space includes both architectural parameters and dataset-level properties. The MLP-specific parameters are the hidden width $\mu_h$, number of layers $\mu_l$, input width $\mu_{x_{\mathrm{in}}}$, activation function $\mu_a$, shuffled categorical ratio $\mu_{p_{\mathrm{shuffle}}}$, node noise scale $\mu_{\mathrm{noise}}$, and layer sparsity $\mu_{\mathrm{sparsity}}$. These are combined with shared dataset parameters controlling the number of observed features $\mu_{z_x}$, number of classes $\mu_{n_c}$, mean number of categories $\mu_{\mathrm{mean\_cat}}$, categorical feature ratio $\mu_{p_{\mathrm{cat}}}$, missing-value ratio $\mu_{p_{\mathrm{missing}}}$, missingness mechanism $\mu_{\mathrm{missingness\_type}}$, class-balance concentration $\mu_{\mathrm{class\_balance}}$, weight scaling $\mu_{\mathrm{weight\_scaling}}$, and input distribution $\mu_{\mathrm{dist}}$. Integer-valued parameters are sampled from truncated normal distributions over the ranges shown in Table \ref{table:params} and then rounded to the nearest integer, while continuous parameters are sampled from truncated normals on their admissible intervals. Categorical parameters are sampled with an $\epsilon$-greedy scheme: with probability $1-\epsilon$ we select the reference value and with probability $\epsilon$ we sample uniformly from the remaining choices. We also implemented Tree-based SCMs, as in related literature, but found that they did not make a significant contribution to our \model\ pipeline, so we exclude them here for brevity \cite{qu_tabicl_2025}.

\begin{table*}[t]
\centering
\caption{MLP-based SCM parameter space and sampling distributions. Numeric parameters are sampled from truncated normal distributions centered at the corresponding mean hyperparameter. Categorical parameters are sampled with an $\epsilon$-greedy rule with $\epsilon = 0.3 \cdot \sigma$. \label{table:params}}
\begin{adjustbox}{max width=1\textwidth}
\begin{tabular}{lccc}
\toprule
\textbf{Parameter} & \textbf{Type} & \textbf{Range} & \textbf{Distribution} \\
\midrule
Mean hidden size $\mu_h$ & integer & [1, 256] & $\text{TruncNorm}(\mu_h, 10 \cdot \sigma)$ \\
Mean number of layers $\mu_l$ & integer & [1, 12] & $\text{TruncNorm}(\mu_l, \sigma)$ \\
Mean input width $\mu_{x_{\mathrm{in}}}$ & integer & [1, 15] & $\text{TruncNorm}(\mu_{x_{\mathrm{in}}}, 2 \cdot \sigma)$ \\
Activation function $\mu_a$ & categorical & \{\texttt{relu}, \texttt{elu}, \texttt{tanh}, \texttt{identity}\} & $\epsilon$-greedy \\
Mean number of features $\mu_{z_x}$ & integer & [2, 200] & $\text{TruncNorm}(\mu_{z_x}, s_{z_x} \cdot \sigma)$ \\
Mean number of classes $\mu_{n_c}$ & integer & [2, 10] & $\text{TruncNorm}(\mu_{n_c}, \sigma)$ \\
Mean number of categories $\mu_{\mathrm{mean\_cat}}$ & integer & [2, 120] & $\text{TruncNorm}(\mu_{\mathrm{mean\_cat}}, 2 \cdot \sigma)$ \\
Categorical feature ratio $\mu_{p_{\mathrm{cat}}}$ & continuous & [0, 1] & $\text{TruncNorm}(\mu_{p_{\mathrm{cat}}}, 0.1 \cdot \sigma)$ \\
Shuffled categorical ratio $\mu_{p_{\mathrm{shuffle}}}$ & continuous & [0, 1] & $\text{TruncNorm}(\mu_{p_{\mathrm{shuffle}}}, 0.1 \cdot \sigma)$ \\
Missing-value ratio $\mu_{p_{\mathrm{missing}}}$ & continuous & [0, 1] & $\text{TruncNorm}(\mu_{p_{\mathrm{missing}}}, 0.1 \cdot \sigma)$ \\
Missingness type $\mu_{\mathrm{missingness\_type}}$ & categorical & \{\texttt{mcar}, \texttt{featurewise}, \texttt{self\_masking}\} & $\epsilon$-greedy \\
Class-balance concentration $\mu_{\mathrm{class\_balance}}$ & continuous & [0.1, 10] & $\text{TruncNorm}(\mu_{\mathrm{class\_balance}}, 1.5 \cdot \sigma)$ \\
Node noise scale $\mu_{\mathrm{noise}}$ & continuous & [0, 1] & $\text{TruncNorm}(\mu_{\mathrm{noise}}, 0.1 \cdot \sigma)$ \\
Layer sparsity $\mu_{\mathrm{sparsity}}$ & continuous & [0, 0.9] & $\text{TruncNorm}(\mu_{\mathrm{sparsity}}, 0.1 \cdot \sigma)$ \\
Weight scaling $\mu_{\mathrm{weight\_scaling}}$ & categorical & \{\texttt{None}, \texttt{Small}, \texttt{Large}\} & $\epsilon$-greedy \\
Input distribution $\mu_{\mathrm{dist}}$ & categorical & \{\texttt{exponential}, \texttt{uniform}, \texttt{normal}\} & $\epsilon$-greedy \\
\bottomrule
\end{tabular}
\end{adjustbox}
\vspace{0.25em}
\end{table*}

\section{Learned SCM Parameter Configurations}\label{app:scm-results}
\textbf{Different Domains Require Different SCM Parameters.} In Table \ref{table:scm-params}, we show a subset of the parameters used to configure the SCM generators, which determine the shape, geometry, and learnability of the generated datasets. For each benchmark, we show the parameters corresponding to the setting with the highest weight, as in Equation \ref{eq:dro-objective}, the optimal response of the adversary when constrained to generate data that is similar to the real-world reference data. In the table, SCM size refers the dimensions of the MLP generating the data, which we split into five categorical values, balance refers to the class imbalance (higher corresponds to greater class balance), and noise refers to the scale of the noise added to the output of each neuron. We can clearly observe the variety in parameterizations across benchmarks, some of which align with what we know about the real-world datasets from that domain. For example, for the Genomics benchmark, most datasets have $d=500$ numeric features, and the domain-adaptation algorithm has correctly constrained the parameter space to datasets with many features (we limit the number of generated features to $d=200$ per dataset). While some parameters, such as the number of features, may be easier to interpret and adjust manually, for others, such as network sparsity or the activation function, it would be difficult to interpret or understand how changing these values would alter the generated datasets. We can see however, both in Table \ref{table:scm-params} and Figure \ref{fig:three-subfigures}, that our domain-adaptation algorithm is not only able to adjust these parameters automatically to align with and cover the real-world datasets in the meta-feature space, but that it is necessary to do so, as their values vary significantly between domains.
\begin{table*}[t]
\centering
\caption{SCM generator parameters with the highest weight in the first iteration of the \model\ algorithm for various benchmarks. Different benchmarks correspond to significantly different data generation regimes. We report only a subset of the parameters in this table.}
\setlength{\tabcolsep}{2pt} % reduce horizontal padding
% \small % smaller font size
\begin{adjustbox}{max width=1\textwidth} % scale table if still too wide
\begin{tabular}{lcccccccccc}
\toprule
 \textbf{Benchmark} &\textbf{SCM Size} &  \textbf{\# Feat.} & \textbf{\# Class} & \textbf{Act.} & \textbf{Sparsity} & $\mathbf{p_{\text{cat}}}$ & \textbf{Cat. Dim.} & $\mathbf{p_{\text{miss}}}$ & \textbf{Balance} & \textbf{Noise}\\
\midrule
\textbf{TabPertNet}\cite{ye_towards_2024}
& Medium & 13 & 2 & tanh & 0.9 & 0.0 & - & 0.4 & 3 & 0.0\\
\midrule
\textbf{TabArena}\cite{erickson_tabarena_2025}
& Medium & 44 & 2 & tanh & 0.9 & 0.5 & 50 & 0.2 & 3 & 0.0\\
\midrule
\textbf{PMLB}\cite{olson2017pmlb}
& Small & 12 & 10 & tanh & 0.9 & 0.0 & - & 0.2 & 3 & 0.0\\
\midrule
\textbf{Engineering} & Large & 65 & 2 & elu & 0.0 & 0.0 & - & 0.0 & 10 & 0.01\\
\midrule
\textbf{Genomics} & Large & 200 & 2 & relu & 0.1 & 0.0 & - & 0.4 & 3 & 0.05\\
\midrule
\textbf{Healthcare} & Small & 5 & 2 & identity & 0.0 & 1.0 & 2 & 0.2 & 10 & 0.5\\
\bottomrule
\end{tabular}
\end{adjustbox}
\label{table:scm-params}
\end{table*}

\section{Generator Coverage on Foreign-Exchange Data}\label{app:forex}
\textbf{Effect of the Generator Family.} To investigate the limitations imposed by the synthetic generator family, we consider a corpus of 64 foreign-exchange datasets from OpenML \cite{bischl_openml_2021}, derived from daily candlestick price series. These datasets exhibit strong temporal dependencies that are not naturally represented by the non-temporal SCMs used in our main experiments. With the original generators, the entropy-regularized optimal transport alignment objective between the real and synthetic dataset embeddings is approximately $S_d=2$, compared with values below $0.2$ on the other evaluated domains. We also observe no empirical coverage ($C=0$), where coverage is the fraction of reference datasets whose nearest synthetic embedding is at least as close as a resampled real-data neighbor, averaged over resampling trials.

To expand the generator family, we introduce a temporal SCM with controllable latent dimension, autoregressive persistence, feature-noise ratio, maximum lag, prediction horizon, and target signal strength. This extended parameter space reduces the alignment objective to approximately $S_d=-0.6$ and increases coverage to $C=0.05$. Although coverage remains limited, the additional temporal structure permits closer alignment with the reference domain. Table \ref{table:forex} reports the resulting mean test AUC. Neither fine-tuning nor \model\ with the original generators improves upon the frozen backbone; in contrast, the temporal extension yields a modest improvement. The results illustrate that reweighting generators cannot compensate for missing data-generating mechanisms, while expanding the SCM family can extend the range of domains amenable to robust, domain-adapted continued training.

\begin{table}[h]
\centering
\caption{Mean test AUC across 64 foreign-exchange datasets, comparing the original and expanded synthetic generator families.}
\setlength{\tabcolsep}{8pt}
\small
\begin{tabular}{lcc}
\toprule
\textbf{Model} & \textbf{Mean AUC} & \textbf{Change vs. base} \\
\midrule
TabPFN V2 & 0.6602 & --- \\
TabPFN V2 (FT) & 0.6551 & $-0.0051$ \\
TabPFN V2 (\model; original SCMs) & 0.6597 & $-0.0005$ \\
TabPFN V2 (\model; temporal SCM) & \textbf{0.6652} & $\mathbf{+0.0050}$ \\
\bottomrule
\end{tabular}
\label{table:forex}
\end{table}

\section{Additional Experiment Details}\label{app:experiments}
In Tables \ref{table:tabpertnet-1}-\ref{table:healthcare} we provide the full experiment results from the TabPertNet \cite{ye_towards_2024}, TabArena \cite{erickson_tabarena_2025}, PMLB \cite{olson2017pmlb}, and domain-specific OpenML \cite{bischl_openml_2021} tabular benchmarks. For all benchmarks, we evaluate on classification datasets only, and limit our evaluation to datasets with at most $n=10k$ samples and $p=500$ features. We use the training data from each dataset during the domain-adaptation process to approximate the reference distribution for the given domain. We report our test results on $N=70$ datasets from TabPertNet, $N=91$ datasets from PMLB, $N=21$ datasets from TabArena, $N=31$ dataset for Healthcare, $N=11$ datasets for Genomics, and $N=20$ datasets for Engineering. The original genomics datasets have $O(10k)$ features, so we first preprocess these datasets by randomly selecting a subset of $d=500$ features and down-sampling. For TabArena, we use the same folds as defined in the original work and corresponding leaderboard. For the OpenML datasets, we use the same train/test splits as provided from the database. For TabPertNet, we use the same train/test split used in the original paper. We report ROC-AUC for binary classification tasks and AUC one-vs-one (OVO) for multi-class tasks.

\textbf{Compute Resources.} All experiments were run on a single 80GB A100 GPU together with up to 140 CPU cores during the robust parameter-search stage. Constructing the domain-adapted reference distribution for a benchmark required approximately 10 minutes of GPU time, while each max-min iteration of the \model\ algorithm required approximately 1 GPU-hour; we used 5 epochs for Genomics, 10 epochs for Engineering and Healthcare, and 30 epochs for TabPertNet, TabArena, and PMLB. The CPU-intensive parameter-search stage used roughly 140 cores for about 6 hours per full training run.

\paragraph{Licensing for existing assets.}
Our experiments use existing open-source benchmarks, datasets, models, and software packages. We cite the original sources for each benchmark and model in the main paper. TabArena is used under its publicly released benchmark resources; its code repository is licensed under Apache License 2.0, and the associated dataset curation repository is licensed under the BSD 3-Clause License. The TabArena authors further state that datasets whose licenses do not allow sharing or use in an academic benchmark are excluded from the benchmark. PMLB is publicly available under the MIT License. OpenML data and metadata are made available under OpenML's terms of use; OpenML states that empirical data and metadata are available under a CC-BY license, while individual dataset metadata include dataset-specific license fields. For TabPertNet, we use the publicly released dataset split from the original benchmark; the TP-BERTa \cite{yan_making_2024} repository distributing the TabPertNet/OpenTabs subset states that the authors obtained permission to distribute the used subset. We use TabPFN-V2 weights under the Apache 2.0 license and the open-source TabICL V2 classifier under the BSD 3-Clause License. Optuna is licensed under the MIT License. We use all assets only for academic research and benchmarking, and we do not redistribute modified versions of these assets.
\begin{table*}[h]
\centering
\caption{TabPertNet \cite{ye_towards_2024} Test Results, Part 1 (AUC $\uparrow$).\label{table:tabpertnet-1}}
\setlength{\tabcolsep}{1.5pt} % reduce horizontal padding
% \small % smaller font size
\begin{adjustbox}{width=1\textwidth} % scale table if still too wide
\begin{tabular}{l*{11}{r}}
\toprule
 & \multicolumn{3}{c}{\textbf{TabPFN V2}} & \multicolumn{3}{c}{\textbf{TabICL V2}} & \multicolumn{5}{c}{\textbf{Conventional baselines}} \\
\cmidrule(lr){2-4}\cmidrule(lr){5-7}\cmidrule(lr){8-12}
\textbf{Dataset} & \textbf{RAD-TFM} & \textbf{Base} & \textbf{FT} & \textbf{RAD-TFM} & \textbf{Base} & \textbf{FT} & \textbf{CatBoost} & \textbf{RF} & \textbf{XGB} & \textbf{LR} & \textbf{MLP} \\
\midrule
BankNoteAuthenticati & 1.000 & 1.000 & 1.000 & 1.000 & 1.000 & 1.000 & 1.000 & 1.000 & 1.000 & 0.999 & 0.998 \\
Bank\_Personal\_Loan\_M & 0.597 & 0.578 & 0.578 & 0.618 & 0.619 & 0.604 & 0.600 & 0.601 & 0.599 & 0.606 & 0.590 \\
Breast\_Cancer & 0.882 & 0.876 & 0.876 & 0.892 & 0.888 & 0.886 & 0.864 & 0.891 & 0.839 & 0.885 & 0.869 \\
Customer\_Behaviour & 0.958 & 0.958 & 0.958 & 0.769 & 0.782 & 0.773 & 0.944 & 0.968 & 0.972 & 0.944 & 0.940 \\
Employee Satisfactio & 0.655 & 0.433 & 0.441 & 0.474 & 0.431 & 0.474 & 0.543 & 0.473 & 0.436 & 0.439 & 0.598 \\
NFL & 0.725 & 0.731 & 0.731 & 0.723 & 0.729 & 0.724 & 0.720 & 0.679 & 0.671 & 0.670 & 0.685 \\
TravelInsurancePredi & 0.708 & 0.670 & 0.668 & 0.716 & 0.688 & 0.701 & 0.708 & 0.546 & 0.631 & 0.715 & 0.723 \\
UniversalBank & 0.597 & 0.578 & 0.578 & 0.618 & 0.619 & 0.604 & 0.600 & 0.593 & 0.599 & 0.606 & 0.590 \\
audit\_data & 0.997 & 0.998 & 0.998 & 1.000 & 1.000 & 1.000 & 1.000 & 1.000 & 1.000 & 0.992 & 0.981 \\
audit\_risk & 0.997 & 0.998 & 0.998 & 1.000 & 1.000 & 1.000 & 1.000 & 1.000 & 1.000 & 0.992 & 0.981 \\
b\_depressed & 0.468 & 0.447 & 0.445 & 0.557 & 0.521 & 0.532 & 0.578 & 0.574 & 0.513 & 0.446 & 0.601 \\
bt\_dataset\_t3 & 1.000 & 0.999 & 0.999 & 0.997 & 0.999 & 0.999 & 0.998 & 0.995 & 0.995 & 0.969 & 0.814 \\
loan & 0.673 & 0.659 & 0.659 & 0.697 & 0.702 & 0.685 & 0.669 & 0.691 & 0.697 & 0.677 & 0.654 \\
mechanical\_analysis & 1.000 & 1.000 & 1.000 & 1.000 & 1.000 & 1.000 & 1.000 & 1.000 & 1.000 & 0.949 & 0.940 \\
new\_model & 1.000 & 1.000 & 1.000 & 1.000 & 1.000 & 1.000 & 1.000 & 1.000 & 1.000 & 1.000 & 1.000 \\
piracydataset & 0.573 & 0.572 & 0.571 & 0.607 & 0.593 & 0.606 & 0.592 & 0.606 & 0.607 & 0.606 & 0.486 \\
0284\_bank8FM & 0.994 & 0.994 & 0.994 & 0.994 & 0.994 & 0.994 & 0.994 & 0.991 & 0.992 & 0.993 & 0.994 \\
0292\_cpu\_small & 0.980 & 0.980 & 0.980 & 0.982 & 0.982 & 0.983 & 0.979 & 0.975 & 0.978 & 0.959 & 0.972 \\
0312\_cpu\_act & 0.987 & 0.988 & 0.988 & 0.989 & 0.988 & 0.988 & 0.984 & 0.983 & 0.986 & 0.970 & 0.982 \\
0345\_delta\_ailerons & 0.977 & 0.975 & 0.975 & 0.978 & 0.978 & 0.978 & 0.974 & 0.975 & 0.972 & 0.967 & 0.971 \\
0356\_delta\_elevators & 0.950 & 0.949 & 0.949 & 0.952 & 0.953 & 0.952 & 0.952 & 0.946 & 0.944 & 0.945 & 0.949 \\
0400\_analcatdata\_sup & 1.000 & 1.000 & 1.000 & 1.000 & 1.000 & 1.000 & 1.000 & 1.000 & 1.000 & 0.938 & 0.750 \\
0419\_pm10 & 0.643 & 0.438 & 0.433 & 0.762 & 0.640 & 0.659 & 0.567 & 0.677 & 0.570 & 0.404 & 0.562 \\
0435\_strikes & 1.000 & 1.000 & 1.000 & 1.000 & 1.000 & 1.000 & 0.996 & 0.990 & 1.000 & 0.616 & 0.555 \\
0437\_quake & 0.521 & 0.432 & 0.431 & 0.564 & 0.490 & 0.566 & 0.447 & 0.500 & 0.482 & 0.563 & 0.531 \\
0445\_arsenic-male-bl & 0.744 & 0.600 & 0.600 & 0.756 & 0.667 & 0.700 & 0.533 & 0.500 & 0.778 & 0.411 & 0.711 \\
0446\_arsenic-female- & 0.858 & 0.845 & 0.844 & 0.837 & 0.839 & 0.845 & 0.807 & 0.737 & 0.786 & 0.814 & 0.663 \\
0472\_analcatdata\_mar & 0.663 & 0.360 & 0.366 & 0.377 & 0.669 & 0.606 & 0.571 & 0.526 & 0.480 & 0.383 & 0.429 \\
0509\_pollen & 0.512 & 0.500 & 0.502 & 0.530 & 0.512 & 0.523 & 0.503 & 0.498 & 0.530 & 0.456 & 0.528 \\
0526\_colleges\_aaup & 1.000 & 1.000 & 1.000 & 0.978 & 0.973 & 0.974 & 1.000 & 1.000 & 0.999 & 0.967 & 0.961 \\
0541\_plasma\_retinol & 0.585 & 0.591 & 0.591 & 0.585 & 0.602 & 0.591 & 0.636 & 0.699 & 0.352 & 0.625 & 0.511 \\
0555\_socmob & 0.992 & 0.987 & 0.987 & 0.948 & 0.959 & 0.955 & 0.995 & 0.989 & 0.992 & 0.927 & 0.931 \\
0885\_compas-two-year & 0.709 & 0.709 & 0.709 & 0.712 & 0.705 & 0.708 & 0.693 & 0.441 & 0.570 & 0.703 & 0.684 \\
0948\_Ishwar & 0.996 & 0.998 & 0.998 & 0.950 & 0.946 & 0.954 & 0.990 & 0.983 & 0.988 & 0.975 & 0.979 \\
1011\_cleve & 0.943 & 0.933 & 0.933 & 0.933 & 0.938 & 0.938 & 0.929 & 0.943 & 0.933 & 0.957 & 0.971 \\
1142\_Sick\_numeric & 0.995 & 0.991 & 0.991 & 0.998 & 0.997 & 0.997 & 0.995 & 0.983 & 0.978 & 0.917 & 0.578 \\
\bottomrule
\end{tabular}
\end{adjustbox}
\end{table*}

\begin{table*}[h]
\centering
\caption{TabPertNet \cite{ye_towards_2024} Test Results, Part 2 (AUC $\uparrow$).\label{table:tabpertnet-2}}
\setlength{\tabcolsep}{1.5pt} % reduce horizontal padding
% \small % smaller font size
\begin{adjustbox}{width=1\textwidth} % scale table if still too wide
\begin{tabular}{l*{11}{r}}
\toprule
 & \multicolumn{3}{c}{\textbf{TabPFN V2}} & \multicolumn{3}{c}{\textbf{TabICL V2}} & \multicolumn{5}{c}{\textbf{Conventional baselines}} \\
\cmidrule(lr){2-4}\cmidrule(lr){5-7}\cmidrule(lr){8-12}
\textbf{Dataset} & \textbf{RAD-TFM} & \textbf{Base} & \textbf{FT} & \textbf{RAD-TFM} & \textbf{Base} & \textbf{FT} & \textbf{CatBoost} & \textbf{RF} & \textbf{XGB} & \textbf{LR} & \textbf{MLP} \\
\midrule
1201\_Gender-Recognit & 1.000 & 1.000 & 1.000 & 0.999 & 0.999 & 0.999 & 0.998 & 0.999 & 0.999 & 0.996 & 0.993 \\
1222\_premier\_league\_ & 0.695 & 0.699 & 0.699 & 0.695 & 0.690 & 0.691 & 0.577 & 0.634 & 0.618 & 0.693 & 0.661 \\
1408\_national-longit & 1.000 & 1.000 & 1.000 & 0.996 & 0.997 & 0.996 & 1.000 & 1.000 & 1.000 & 0.992 & 0.990 \\
1413\_shill-bidding & 1.000 & 1.000 & 1.000 & 1.000 & 1.000 & 1.000 & 1.000 & 1.000 & 1.000 & 0.996 & 0.997 \\
1458\_kdd\_ipums\_la\_97 & 0.938 & 0.937 & 0.937 & 0.938 & 0.939 & 0.936 & 0.939 & 0.932 & 0.923 & 0.890 & 0.909 \\
1461\_heart-failure & 0.861 & 0.861 & 0.861 & 0.861 & 0.878 & 0.894 & 0.822 & 0.844 & 0.856 & 0.944 & 0.794 \\
1512\_eye\_movements & 0.714 & 0.746 & 0.745 & 0.818 & 0.747 & 0.817 & 0.690 & 0.698 & 0.707 & 0.594 & 0.627 \\
1564\_Mammographic-Ma & 0.817 & 0.815 & 0.815 & 0.818 & 0.818 & 0.815 & 0.808 & 0.701 & 0.658 & 0.807 & 0.691 \\
1578\_kdd\_ipums\_la\_97 & 0.954 & 0.953 & 0.953 & 0.956 & 0.954 & 0.955 & 0.959 & 0.954 & 0.946 & 0.907 & 0.927 \\
1592\_Diabetes-Data-S & 0.873 & 0.871 & 0.871 & 0.877 & 0.873 & 0.871 & 0.876 & 0.896 & 0.871 & 0.839 & 0.772 \\
1600\_VulNoneVul & 0.801 & 0.798 & 0.797 & 0.757 & 0.762 & 0.758 & 0.720 & 0.638 & 0.698 & 0.755 & 0.523 \\
1635\_Is-this-a-good- & 0.810 & 0.805 & 0.804 & 0.779 & 0.787 & 0.781 & 0.820 & 0.779 & 0.752 & 0.747 & 0.750 \\
1692\_Gender-Classifi & 0.995 & 0.995 & 0.995 & 0.993 & 0.993 & 0.994 & 0.993 & 0.992 & 0.989 & 0.993 & 0.993 \\
1736\_combined-wine-d & 1.000 & 1.000 & 1.000 & 1.000 & 1.000 & 1.000 & 1.000 & 1.000 & 1.000 & 0.999 & 0.998 \\
1742\_Loan-Predicatio & 0.785 & 0.743 & 0.743 & 0.712 & 0.733 & 0.728 & 0.766 & 0.769 & 0.737 & 0.742 & 0.752 \\
1752\_Wisconsin-breas & 0.996 & 0.994 & 0.994 & 0.992 & 0.992 & 0.992 & 0.990 & 0.992 & 0.987 & 0.992 & 0.977 \\
1759\_Red\_White-wine- & 1.000 & 1.000 & 1.000 & 1.000 & 1.000 & 1.000 & 1.000 & 1.000 & 1.000 & 0.999 & 0.998 \\
1898\_Personal-Loan-M & 0.618 & 0.624 & 0.623 & 0.615 & 0.619 & 0.616 & 0.631 & 0.645 & 0.603 & 0.627 & 0.587 \\
2304\_electricity\_see & 0.875 & 0.866 & 0.866 & 0.889 & 0.887 & 0.884 & 0.853 & 0.867 & 0.875 & 0.699 & 0.770 \\
2305\_electricity\_see & 0.934 & 0.937 & 0.937 & 0.915 & 0.913 & 0.919 & 0.905 & 0.917 & 0.922 & 0.788 & 0.858 \\
2306\_electricity\_see & 0.877 & 0.879 & 0.879 & 0.863 & 0.858 & 0.862 & 0.862 & 0.874 & 0.871 & 0.771 & 0.827 \\
2308\_electricity\_see & 0.912 & 0.911 & 0.911 & 0.903 & 0.902 & 0.901 & 0.891 & 0.886 & 0.886 & 0.717 & 0.837 \\
2389\_airlines\_seed\_0 & 0.679 & 0.683 & 0.683 & 0.650 & 0.644 & 0.656 & 0.640 & 0.618 & 0.637 & 0.683 & 0.666 \\
2390\_airlines\_seed\_1 & 0.646 & 0.640 & 0.640 & 0.643 & 0.638 & 0.650 & 0.637 & 0.607 & 0.585 & 0.662 & 0.646 \\
2391\_airlines\_seed\_2 & 0.650 & 0.641 & 0.641 & 0.636 & 0.629 & 0.632 & 0.619 & 0.618 & 0.650 & 0.647 & 0.635 \\
2392\_airlines\_seed\_3 & 0.607 & 0.623 & 0.623 & 0.583 & 0.585 & 0.580 & 0.607 & 0.600 & 0.564 & 0.612 & 0.617 \\
2393\_airlines\_seed\_4 & 0.677 & 0.680 & 0.679 & 0.686 & 0.686 & 0.686 & 0.647 & 0.639 & 0.607 & 0.690 & 0.682 \\
2619\_sf-police-incid & 0.465 & 0.450 & 0.450 & 0.596 & 0.604 & 0.601 & 0.530 & 0.595 & 0.410 & 0.495 & 0.445 \\
2620\_sf-police-incid & 0.509 & 0.487 & 0.485 & 0.464 & 0.456 & 0.455 & 0.518 & 0.510 & 0.457 & 0.512 & 0.456 \\
2621\_sf-police-incid & 0.478 & 0.540 & 0.544 & 0.618 & 0.570 & 0.569 & 0.731 & 0.739 & 0.616 & 0.646 & 0.449 \\
2622\_sf-police-incid & 0.522 & 0.458 & 0.458 & 0.623 & 0.386 & 0.361 & 0.482 & 0.442 & 0.430 & 0.449 & 0.467 \\
2703\_compas-two-year & 0.729 & 0.730 & 0.730 & 0.733 & 0.729 & 0.733 & 0.711 & 0.525 & 0.585 & 0.732 & 0.733 \\
trial & 1.000 & 1.000 & 1.000 & 1.000 & 1.000 & 1.000 & 1.000 & 1.000 & 1.000 & 1.000 & 1.000 \\
wines\_SPA & 1.000 & 1.000 & 1.000 & 1.000 & 1.000 & 1.000 & 1.000 & 1.000 & 1.000 & 1.000 & 1.000 \\
\bottomrule
\end{tabular}
\end{adjustbox}
\end{table*}

\begin{table*}[h]
\centering
\caption{TabArena \cite{erickson_tabarena_2025} Test Results (OVO AUC $\uparrow$).\label{table:tabarena}}
\setlength{\tabcolsep}{1.5pt} % reduce horizontal padding
% \small % smaller font size
\begin{adjustbox}{width=1\textwidth} % scale table if still too wide
\begin{tabular}{l*{11}{r}}
\toprule
 & \multicolumn{3}{c}{\textbf{TabPFN V2}} & \multicolumn{3}{c}{\textbf{TabICL V2}} & \multicolumn{5}{c}{\textbf{Conventional baselines}} \\
\cmidrule(lr){2-4}\cmidrule(lr){5-7}\cmidrule(lr){8-12}
\textbf{Dataset} & \textbf{RAD-TFM} & \textbf{Base} & \textbf{FT} & \textbf{RAD-TFM} & \textbf{Base} & \textbf{FT} & \textbf{CatBoost} & \textbf{RF} & \textbf{XGB} & \textbf{LR} & \textbf{MLP} \\
\midrule
Bank\_Customer\_Churn & 0.876 & 0.875 & 0.874 & 0.841 & 0.825 & 0.802 & 0.878 & 0.845 & 0.845 & 0.745 & 0.848 \\
Fitness\_Club & 0.811 & 0.810 & 0.810 & 0.779 & 0.778 & 0.777 & 0.803 & 0.760 & 0.757 & 0.796 & 0.563 \\
Is-this-a-good-custo & 0.731 & 0.707 & 0.712 & 0.752 & 0.749 & 0.746 & 0.729 & 0.655 & 0.670 & 0.712 & 0.585 \\
MIC & 0.643 & 0.625 & 0.634 & 0.674 & 0.684 & 0.678 & 0.684 & 0.678 & 0.652 & 0.722 & 0.635 \\
Marketing\_Campaign & 0.908 & 0.897 & 0.894 & 0.850 & 0.830 & 0.793 & 0.907 & 0.894 & 0.909 & 0.837 & 0.647 \\
NATICUSdroid & 0.986 & 0.984 & 0.985 & 0.987 & 0.987 & 0.987 & 0.985 & 0.978 & 0.984 & 0.982 & 0.981 \\
anneal & 0.992 & 1.000 & 1.000 & 0.999 & 1.000 & 1.000 & 0.958 & 0.999 & 0.994 & 0.953 & 0.961 \\
blood-transfusion-se & 0.741 & 0.729 & 0.732 & 0.741 & 0.740 & 0.744 & 0.697 & 0.663 & 0.649 & 0.744 & 0.723 \\
churn & 0.944 & 0.940 & 0.942 & 0.942 & 0.944 & 0.946 & 0.938 & 0.927 & 0.933 & 0.814 & 0.882 \\
coil2000\_insurance\_p & 0.752 & 0.746 & 0.749 & 0.765 & 0.758 & 0.762 & 0.721 & 0.689 & 0.702 & 0.735 & 0.696 \\
credit-g & 0.780 & 0.781 & 0.782 & 0.772 & 0.772 & 0.774 & 0.774 & 0.771 & 0.772 & 0.773 & 0.704 \\
diabetes & 0.841 & 0.842 & 0.844 & 0.845 & 0.847 & 0.845 & 0.830 & 0.828 & 0.808 & 0.839 & 0.781 \\
hazelnut-spread-cont & 0.982 & 0.982 & 0.983 & 0.992 & 0.992 & 0.992 & 0.959 & 0.950 & 0.968 & 0.584 & 0.931 \\
maternal\_health\_risk & 0.962 & 0.952 & 0.952 & 0.964 & 0.966 & 0.968 & 0.952 & 0.955 & 0.958 & 0.804 & 0.793 \\
polish\_companies\_ban & 0.967 & 0.963 & 0.966 & 0.980 & 0.980 & 0.982 & 0.970 & 0.945 & 0.963 & 0.532 & 0.814 \\
qsar-biodeg & 0.930 & 0.925 & 0.930 & 0.933 & 0.935 & 0.934 & 0.925 & 0.915 & 0.920 & 0.921 & 0.919 \\
seismic-bumps & 0.773 & 0.782 & 0.778 & 0.763 & 0.754 & 0.762 & 0.758 & 0.761 & 0.740 & 0.766 & 0.602 \\
splice & 0.998 & 0.997 & 0.998 & 0.998 & 0.997 & 0.998 & 0.996 & 0.997 & 0.995 & 0.992 & 0.988 \\
students\_dropout\_and & 0.869 & 0.864 & 0.866 & 0.862 & 0.851 & 0.860 & 0.857 & 0.856 & 0.869 & 0.864 & 0.832 \\
taiwanese\_bankruptcy & 0.948 & 0.948 & 0.950 & 0.946 & 0.949 & 0.948 & 0.941 & 0.941 & 0.940 & 0.592 & 0.832 \\
website\_phishing & 0.973 & 0.973 & 0.973 & 0.978 & 0.978 & 0.979 & 0.976 & 0.970 & 0.972 & 0.926 & 0.869 \\
\bottomrule
\end{tabular}
\end{adjustbox}
\end{table*}

\begin{table*}[h]
\centering
\caption{PMLB \cite{olson2017pmlb} Test Results, Part 1 (OVO AUC $\uparrow$).\label{table:pmlb-1}}
\setlength{\tabcolsep}{1.5pt} % reduce horizontal padding
% \small % smaller font size
\begin{adjustbox}{width=1\textwidth} % scale table if still too wide
\begin{tabular}{l*{11}{r}}
\toprule
 & \multicolumn{3}{c}{\textbf{TabPFN V2}} & \multicolumn{3}{c}{\textbf{TabICL V2}} & \multicolumn{5}{c}{\textbf{Conventional baselines}} \\
\cmidrule(lr){2-4}\cmidrule(lr){5-7}\cmidrule(lr){8-12}
\textbf{Dataset} & \textbf{RAD-TFM} & \textbf{Base} & \textbf{FT} & \textbf{RAD-TFM} & \textbf{Base} & \textbf{FT} & \textbf{CatBoost} & \textbf{RF} & \textbf{XGB} & \textbf{LR} & \textbf{MLP} \\
\midrule
1028\_SWD & 0.841 & 0.843 & 0.843 & 0.861 & 0.868 & 0.864 & 0.861 & 0.861 & 0.861 & 0.860 & 0.827 \\
1029\_LEV & 0.902 & 0.901 & 0.902 & 0.904 & 0.915 & 0.909 & 0.917 & 0.911 & 0.916 & 0.885 & 0.849 \\
1030\_ERA & 0.788 & 0.787 & 0.787 & 0.785 & 0.782 & 0.782 & 0.742 & 0.737 & 0.739 & 0.788 & 0.724 \\
294\_satellite\_image & 0.995 & 0.995 & 0.995 & 0.996 & 0.996 & 0.996 & 0.989 & 0.988 & 0.991 & 0.957 & 0.985 \\
GAMETES\_Epistasis\_2\_ & 0.739 & 0.738 & 0.737 & 0.740 & 0.732 & 0.694 & 0.684 & 0.601 & 0.662 & 0.530 & 0.464 \\
GAMETES\_Epistasis\_2\_ & 0.876 & 0.871 & 0.872 & 0.883 & 0.876 & 0.897 & 0.851 & 0.711 & 0.807 & 0.498 & 0.666 \\
GAMETES\_Epistasis\_3\_ & 0.480 & 0.482 & 0.482 & 0.704 & 0.448 & 0.690 & 0.619 & 0.563 & 0.617 & 0.539 & 0.465 \\
GAMETES\_Heterogeneit & 0.731 & 0.727 & 0.727 & 0.737 & 0.689 & 0.717 & 0.768 & 0.725 & 0.769 & 0.533 & 0.637 \\
GAMETES\_Heterogeneit & 0.795 & 0.783 & 0.785 & 0.789 & 0.777 & 0.812 & 0.745 & 0.620 & 0.668 & 0.513 & 0.535 \\
Hill\_Valley\_with\_noi & 0.532 & 0.511 & 0.510 & 0.490 & 0.465 & 0.532 & 0.486 & 0.558 & 0.467 & 0.496 & 0.502 \\
Hill\_Valley\_without\_ & 0.542 & 0.474 & 0.474 & 0.542 & 0.495 & 0.462 & 0.477 & 0.587 & 0.458 & 0.496 & 0.495 \\
agaricus\_lepiota & 1.000 & 1.000 & 1.000 & 1.000 & 1.000 & 1.000 & 1.000 & 1.000 & 1.000 & 0.975 & 1.000 \\
allbp & 0.995 & 0.995 & 0.995 & 0.997 & 0.996 & 0.996 & 0.991 & 0.994 & 0.993 & 0.932 & 0.659 \\
allhypo & 0.986 & 0.984 & 0.984 & 0.991 & 0.994 & 0.993 & 0.982 & 0.983 & 0.981 & 0.840 & 0.707 \\
allrep & 0.904 & 0.899 & 0.899 & 0.926 & 0.910 & 0.915 & 0.898 & 0.929 & 0.890 & 0.821 & 0.456 \\
analcatdata\_authorsh & 1.000 & 1.000 & 1.000 & 1.000 & 1.000 & 1.000 & 1.000 & 1.000 & 1.000 & 0.999 & 0.999 \\
analcatdata\_dmft & 0.621 & 0.605 & 0.614 & 0.602 & 0.620 & 0.603 & 0.533 & 0.517 & 0.548 & 0.597 & 0.449 \\
analcatdata\_germangs & 0.839 & 0.835 & 0.837 & 0.941 & 0.853 & 0.957 & 0.660 & 0.595 & 0.660 & 0.698 & 0.573 \\
analcatdata\_lawsuit & 1.000 & 1.000 & 1.000 & 1.000 & 1.000 & 1.000 & 1.000 & 1.000 & 1.000 & 1.000 & 0.987 \\
ann\_thyroid & 0.998 & 0.999 & 0.999 & 1.000 & 1.000 & 1.000 & 1.000 & 1.000 & 1.000 & 0.789 & 0.922 \\
balance\_scale & 0.985 & 0.986 & 0.986 & 1.000 & 1.000 & 1.000 & 0.915 & 0.754 & 0.871 & 0.941 & 0.852 \\
banana & 0.967 & 0.967 & 0.967 & 0.967 & 0.967 & 0.968 & 0.966 & 0.960 & 0.958 & 0.501 & 0.945 \\
breast\_cancer & 0.800 & 0.774 & 0.774 & 0.852 & 0.783 & 0.791 & 0.800 & 0.783 & 0.539 & 0.800 & 0.626 \\
breast\_cancer\_wiscon & 1.000 & 1.000 & 1.000 & 1.000 & 1.000 & 1.000 & 1.000 & 1.000 & 0.997 & 1.000 & 0.929 \\
breast\_cancer\_wiscon & 0.985 & 0.986 & 0.986 & 0.986 & 0.986 & 0.988 & 0.992 & 0.987 & 0.984 & 0.986 & 0.997 \\
bupa & 0.671 & 0.667 & 0.667 & 0.746 & 0.658 & 0.679 & 0.650 & 0.625 & 0.496 & 0.688 & 0.667 \\
calendarDOW & 0.855 & 0.848 & 0.846 & 0.838 & 0.869 & 0.853 & 0.836 & 0.861 & 0.843 & 0.801 & 0.779 \\
car\_evaluation & 0.999 & 0.999 & 0.999 & 1.000 & 1.000 & 1.000 & 1.000 & 0.999 & 1.000 & 0.921 & 0.862 \\
cars & 1.000 & 1.000 & 1.000 & 1.000 & 1.000 & 1.000 & 0.931 & 0.932 & 0.929 & 0.781 & 0.628 \\
chess & 0.999 & 0.999 & 0.999 & 1.000 & 1.000 & 1.000 & 1.000 & 1.000 & 0.999 & 0.996 & 0.998 \\
churn & 0.974 & 0.973 & 0.973 & 0.974 & 0.969 & 0.961 & 0.964 & 0.960 & 0.969 & 0.769 & 0.580 \\
clean1 & 1.000 & 1.000 & 1.000 & 1.000 & 1.000 & 1.000 & 1.000 & 1.000 & 1.000 & 1.000 & 0.918 \\
clean2 & 1.000 & 1.000 & 1.000 & 1.000 & 1.000 & 1.000 & 1.000 & 1.000 & 1.000 & 1.000 & 1.000 \\
coil2000 & 0.769 & 0.763 & 0.763 & 0.759 & 0.751 & 0.761 & 0.742 & 0.656 & 0.741 & 0.723 & 0.601 \\
congressional\_voting & 1.000 & 1.000 & 1.000 & 1.000 & 1.000 & 1.000 & 1.000 & 1.000 & 0.981 & 0.998 & 0.976 \\
contraceptive\_method & 0.729 & 0.730 & 0.730 & 0.715 & 0.724 & 0.712 & 0.688 & 0.701 & 0.702 & 0.670 & 0.648 \\
credit\_approval\_aust & 0.959 & 0.958 & 0.958 & 0.965 & 0.962 & 0.964 & 0.960 & 0.966 & 0.944 & 0.951 & 0.918 \\
credit\_approval\_germ & 0.844 & 0.846 & 0.846 & 0.827 & 0.838 & 0.821 & 0.829 & 0.798 & 0.733 & 0.824 & 0.735 \\
\bottomrule
\end{tabular}
\end{adjustbox}
\end{table*}

\begin{table*}[h]
\centering
\caption{PMLB \cite{olson2017pmlb} Test Results, Part 2 (OVO AUC $\uparrow$).\label{table:pmlb-2}}
\setlength{\tabcolsep}{1.5pt} % reduce horizontal padding
% \small % smaller font size
\begin{adjustbox}{width=1\textwidth} % scale table if still too wide
\begin{tabular}{l*{11}{r}}
\toprule
 & \multicolumn{3}{c}{\textbf{TabPFN V2}} & \multicolumn{3}{c}{\textbf{TabICL V2}} & \multicolumn{5}{c}{\textbf{Conventional baselines}} \\
\cmidrule(lr){2-4}\cmidrule(lr){5-7}\cmidrule(lr){8-12}
\textbf{Dataset} & \textbf{RAD-TFM} & \textbf{Base} & \textbf{FT} & \textbf{RAD-TFM} & \textbf{Base} & \textbf{FT} & \textbf{CatBoost} & \textbf{RF} & \textbf{XGB} & \textbf{LR} & \textbf{MLP} \\
\midrule
dis & 0.999 & 1.000 & 1.000 & 0.999 & 0.997 & 0.998 & 0.996 & 0.995 & 0.998 & 0.732 & 0.771 \\
dna & 0.995 & 0.995 & 0.995 & 0.997 & 0.997 & 0.997 & 0.997 & 0.996 & 0.998 & 0.990 & 0.991 \\
ecoli & 0.960 & 0.960 & 0.960 & 0.960 & 0.959 & 0.960 & 0.955 & 0.956 & 0.947 & 0.941 & 0.872 \\
haberman & 0.725 & 0.683 & 0.683 & 0.692 & 0.658 & 0.675 & 0.608 & 0.529 & 0.375 & 0.717 & 0.875 \\
heart\_disease\_clevel & 0.904 & 0.894 & 0.894 & 0.904 & 0.909 & 0.918 & 0.904 & 0.916 & 0.904 & 0.904 & 0.880 \\
heart\_disease\_hungar & 0.911 & 0.900 & 0.900 & 0.932 & 0.963 & 0.974 & 0.947 & 0.945 & 0.895 & 0.905 & 0.574 \\
horse\_colic\_outcome & 0.764 & 0.749 & 0.753 & 0.865 & 0.855 & 0.861 & 0.812 & 0.828 & 0.792 & 0.731 & 0.776 \\
horse\_colic\_surgery & 0.821 & 0.810 & 0.810 & 0.786 & 0.806 & 0.821 & 0.790 & 0.804 & 0.722 & 0.841 & 0.587 \\
hypothyroid & 0.967 & 0.968 & 0.968 & 0.973 & 0.980 & 0.978 & 0.979 & 0.979 & 0.977 & 0.865 & 0.546 \\
ionosphere & 1.000 & 0.996 & 0.996 & 0.996 & 1.000 & 1.000 & 1.000 & 1.000 & 1.000 & 0.930 & 0.816 \\
irish & 1.000 & 1.000 & 1.000 & 1.000 & 1.000 & 1.000 & 1.000 & 1.000 & 1.000 & 0.869 & 0.902 \\
kr\_vs\_kp & 1.000 & 1.000 & 1.000 & 1.000 & 1.000 & 1.000 & 1.000 & 1.000 & 1.000 & 0.993 & 0.998 \\
led24 & 0.948 & 0.948 & 0.948 & 0.951 & 0.952 & 0.951 & 0.941 & 0.942 & 0.929 & 0.948 & 0.946 \\
led7 & 0.957 & 0.957 & 0.957 & 0.958 & 0.960 & 0.959 & 0.949 & 0.947 & 0.952 & 0.958 & 0.955 \\
mfeat\_factors & 1.000 & 1.000 & 1.000 & 1.000 & 1.000 & 1.000 & 0.999 & 1.000 & 1.000 & 1.000 & 1.000 \\
mfeat\_fourier & 0.992 & 0.993 & 0.993 & 0.996 & 0.996 & 0.997 & 0.986 & 0.986 & 0.988 & 0.981 & 0.977 \\
mfeat\_karhunen & 1.000 & 1.000 & 1.000 & 1.000 & 1.000 & 1.000 & 0.999 & 0.999 & 0.999 & 1.000 & 0.999 \\
mfeat\_morphological & 0.968 & 0.969 & 0.969 & 0.979 & 0.969 & 0.978 & 0.962 & 0.955 & 0.958 & 0.960 & 0.954 \\
mfeat\_pixel & 1.000 & 1.000 & 1.000 & 1.000 & 1.000 & 1.000 & 1.000 & 1.000 & 1.000 & 1.000 & 0.999 \\
mfeat\_zernike & 0.990 & 0.990 & 0.990 & 0.995 & 0.994 & 0.997 & 0.974 & 0.973 & 0.966 & 0.982 & 0.983 \\
mofn\_3\_7\_10 & 1.000 & 1.000 & 1.000 & 1.000 & 1.000 & 1.000 & 1.000 & 1.000 & 1.000 & 1.000 & 1.000 \\
monk1 & 1.000 & 1.000 & 1.000 & 1.000 & 1.000 & 1.000 & 1.000 & 1.000 & 1.000 & 0.374 & 0.760 \\
monk2 & 1.000 & 1.000 & 1.000 & 1.000 & 1.000 & 1.000 & 1.000 & 0.968 & 1.000 & 0.559 & 0.544 \\
monk3 & 1.000 & 1.000 & 1.000 & 1.000 & 1.000 & 1.000 & 1.000 & 0.998 & 1.000 & 0.871 & 0.871 \\
mushroom & 1.000 & 1.000 & 1.000 & 1.000 & 1.000 & 1.000 & 1.000 & 1.000 & 1.000 & 0.977 & 1.000 \\
optdigits & 1.000 & 1.000 & 1.000 & 1.000 & 1.000 & 1.000 & 0.999 & 1.000 & 0.999 & 0.999 & 0.999 \\
page\_blocks & 0.963 & 0.963 & 0.963 & 0.963 & 0.961 & 0.973 & 0.964 & 0.966 & 0.962 & 0.903 & 0.962 \\
parity5+5 & 1.000 & 1.000 & 1.000 & 1.000 & 0.733 & 1.000 & 0.531 & 0.651 & 1.000 & 0.643 & 0.333 \\
pendigits & 1.000 & 1.000 & 1.000 & 1.000 & 1.000 & 1.000 & 1.000 & 1.000 & 1.000 & 0.996 & 1.000 \\
penguins & 1.000 & 1.000 & 1.000 & 1.000 & 1.000 & 1.000 & 1.000 & 1.000 & 1.000 & 1.000 & 0.901 \\
phoneme & 0.951 & 0.954 & 0.954 & 0.973 & 0.968 & 0.971 & 0.938 & 0.953 & 0.946 & 0.835 & 0.887 \\
profb & 0.698 & 0.670 & 0.671 & 0.700 & 0.710 & 0.580 & 0.652 & 0.671 & 0.680 & 0.656 & 0.579 \\
ring & 0.996 & 0.996 & 0.996 & 0.997 & 0.997 & 0.997 & 0.995 & 0.994 & 0.995 & 0.837 & 0.993 \\
saheart & 0.767 & 0.753 & 0.757 & 0.755 & 0.776 & 0.765 & 0.809 & 0.739 & 0.712 & 0.772 & 0.658 \\
satimage & 0.995 & 0.995 & 0.995 & 0.996 & 0.996 & 0.996 & 0.989 & 0.988 & 0.991 & 0.957 & 0.985 \\
schizo & 0.578 & 0.567 & 0.567 & 0.534 & 0.459 & 0.469 & 0.444 & 0.421 & 0.443 & 0.583 & 0.533 \\
segmentation & 1.000 & 1.000 & 1.000 & 1.000 & 1.000 & 1.000 & 1.000 & 1.000 & 1.000 & 0.995 & 0.994 \\
spambase & 0.988 & 0.989 & 0.989 & 0.992 & 0.993 & 0.993 & 0.988 & 0.980 & 0.990 & 0.976 & 0.971 \\
spect & 0.881 & 0.887 & 0.887 & 0.925 & 0.887 & 0.900 & 0.862 & 0.875 & 0.881 & 0.919 & 0.875 \\
spectf & 0.914 & 0.919 & 0.919 & 0.924 & 0.955 & 0.944 & 0.949 & 0.944 & 0.899 & 0.848 & 0.828 \\
splice & 0.997 & 0.997 & 0.997 & 0.999 & 0.999 & 0.999 & 0.998 & 0.995 & 0.998 & 0.944 & 0.956 \\
threeOf9 & 1.000 & 1.000 & 1.000 & 1.000 & 1.000 & 1.000 & 1.000 & 1.000 & 1.000 & 0.961 & 0.969 \\
tic\_tac\_toe & 0.998 & 0.999 & 0.999 & 1.000 & 0.998 & 0.999 & 0.999 & 0.995 & 1.000 & 0.602 & 0.778 \\
titanic & 0.805 & 0.804 & 0.804 & 0.791 & 0.787 & 0.789 & 0.783 & 0.728 & 0.742 & 0.713 & 0.705 \\
tokyo1 & 0.980 & 0.977 & 0.977 & 0.981 & 0.982 & 0.981 & 0.987 & 0.984 & 0.984 & 0.883 & 0.951 \\
twonorm & 0.998 & 0.998 & 0.998 & 0.998 & 0.998 & 0.998 & 0.998 & 0.997 & 0.997 & 0.998 & 0.998 \\
vehicle & 0.978 & 0.978 & 0.977 & 0.992 & 0.984 & 0.987 & 0.941 & 0.949 & 0.938 & 0.941 & 0.818 \\
waveform\_21 & 0.972 & 0.972 & 0.972 & 0.973 & 0.973 & 0.972 & 0.964 & 0.966 & 0.966 & 0.972 & 0.967 \\
waveform\_40 & 0.972 & 0.973 & 0.973 & 0.973 & 0.974 & 0.973 & 0.960 & 0.967 & 0.968 & 0.966 & 0.964 \\
wine\_quality\_red & 0.814 & 0.822 & 0.822 & 0.852 & 0.859 & 0.856 & 0.821 & 0.871 & 0.849 & 0.777 & 0.673 \\
wine\_quality\_white & 0.860 & 0.876 & 0.875 & 0.895 & 0.881 & 0.884 & 0.783 & 0.799 & 0.809 & 0.639 & 0.635 \\
xd6 & 1.000 & 1.000 & 1.000 & 1.000 & 1.000 & 1.000 & 1.000 & 1.000 & 1.000 & 0.886 & 0.996 \\
yeast & 0.873 & 0.876 & 0.875 & 0.874 & 0.888 & 0.882 & 0.853 & 0.870 & 0.870 & 0.827 & 0.814 \\
\bottomrule
\end{tabular}
\end{adjustbox}
\end{table*}

\begin{table*}[h]
\centering
\caption{Engineering \cite{bischl_openml_2021} Test Results (OVO AUC $\uparrow$).\label{table:engineering}}
\setlength{\tabcolsep}{2pt} % reduce horizontal padding
% \small % smaller font size
\begin{adjustbox}{width=1\textwidth} % scale table if still too wide
\begin{tabular}{ccccccccc}
\toprule
 \textbf{Dataset} & \textbf{TabPFN\textsubscript{n.e.} (RAD-TFM)} & \textbf{TabPFN\textsubscript{n.e.} (Base)} & \textbf{TabPFN\textsubscript{n.e.} (Finetuned)} & \textbf{CatBoost} & \textbf{RF} & \textbf{XGB} & \textbf{LR} & \textbf{MLP} \\
\midrule
anneal\_2 & 1.000 & 1.000 & 1.000 & 1.000 & 1.000 & 1.000 & 0.970 & 0.997 \\
cnae-9\_1468 & 0.999 & 0.998 & 0.998 & 0.998 & 0.997 & 0.997 & 0.998 & 0.993 \\
glass\_41 & 0.975 & 0.964 & 0.964 & 0.952 & 0.939 & 0.937 & 0.819 & 0.497 \\
kc1\_1067 & 0.868 & 0.861 & 0.861 & 0.840 & 0.849 & 0.802 & 0.789 & 0.820 \\
kc2\_1063 & 0.868 & 0.864 & 0.864 & 0.892 & 0.868 & 0.835 & 0.712 & 0.831 \\
kc3\_1065 & 0.780 & 0.774 & 0.774 & 0.750 & 0.848 & 0.685 & 0.554 & 0.702 \\
mc1\_1056 & 0.912 & 0.922 & 0.921 & 0.990 & 0.922 & 0.992 & 0.917 & 0.884 \\
mw1\_1071 & 0.842 & 0.868 & 0.868 & 0.719 & 0.711 & 0.588 & 0.833 & 0.640 \\
pc1\_1068 & 0.882 & 0.875 & 0.876 & 0.875 & 0.805 & 0.821 & 0.692 & 0.533 \\
pc2\_1069 & 0.990 & 0.990 & 0.990 & 0.993 & 0.982 & 0.985 & 0.035 & 0.250 \\
pc3\_1050 & 0.847 & 0.825 & 0.825 & 0.785 & 0.809 & 0.731 & 0.640 & 0.719 \\
pc4\_1049 & 0.946 & 0.930 & 0.930 & 0.938 & 0.934 & 0.949 & 0.918 & 0.880 \\
seismic-bumps\_1500 & 0.993 & 0.993 & 0.993 & 0.966 & 0.966 & 0.980 & 0.993 & 0.748 \\
semeion\_1501 & 0.998 & 0.996 & 0.996 & 0.997 & 0.996 & 0.995 & 0.995 & 0.991 \\
sonar\_40 & 0.963 & 0.963 & 0.963 & 0.981 & 0.954 & 0.972 & 0.926 & 0.843 \\
steel-plates-fault\_1 & 1.000 & 1.000 & 1.000 & 1.000 & 1.000 & 1.000 & 0.698 & 1.000 \\
steel-plates-fault\_4 & 0.980 & 0.980 & 0.980 & 0.974 & 0.973 & 0.971 & 0.737 & 0.950 \\
wall-robot-navigatio & 1.000 & 1.000 & 1.000 & 1.000 & 1.000 & 1.000 & 0.908 & 0.986 \\
wall-robot-navigatio & 1.000 & 1.000 & 1.000 & 1.000 & 1.000 & 1.000 & 0.986 & 1.000 \\
wall-robot-navigatio & 1.000 & 1.000 & 1.000 & 1.000 & 1.000 & 1.000 & 0.987 & 0.999 \\
\bottomrule
\end{tabular}
\end{adjustbox}
\end{table*}

\begin{table*}[h]
\centering
\caption{Genomics \cite{bischl_openml_2021} Test Results (AUC $\uparrow$).\label{table:genomics}}
\setlength{\tabcolsep}{2pt} % reduce horizontal padding
% \small % smaller font size
\begin{adjustbox}{width=1\textwidth} % scale table if still too wide
\begin{tabular}{ccccccccc}
\toprule
 \textbf{Dataset} & \textbf{TabPFN\textsubscript{n.e.} (RAD-TFM)} & \textbf{TabPFN\textsubscript{n.e.} (Base)} & \textbf{TabPFN\textsubscript{n.e.} (Finetuned)} & \textbf{CatBoost} & \textbf{RF} & \textbf{XGB} & \textbf{LR} & \textbf{MLP} \\
\midrule
AP\_Breast\_Omentum\_11 & 0.914 & 0.914 & 0.914 & 0.882 & 0.882 & 0.879 & 0.866 & 0.864 \\
AP\_Colon\_Omentum\_114 & 1.000 & 1.000 & 1.000 & 0.996 & 0.996 & 1.000 & 0.970 & 1.000 \\
AP\_Colon\_Ovary\_1153 & 0.984 & 0.986 & 0.986 & 0.984 & 0.974 & 0.972 & 0.948 & 0.950 \\
AP\_Colon\_Uterus\_1160 & 0.970 & 0.962 & 0.962 & 0.970 & 0.929 & 0.951 & 0.967 & 0.967 \\
AP\_Omentum\_Kidney\_11 & 0.990 & 0.990 & 0.990 & 0.995 & 0.990 & 0.995 & 0.971 & 0.971 \\
AP\_Omentum\_Lung\_1132 & 0.990 & 0.990 & 0.990 & 0.981 & 0.990 & 0.962 & 0.962 & 0.971 \\
AP\_Omentum\_Ovary\_115 & 0.975 & 0.956 & 0.956 & 0.975 & 0.944 & 0.944 & 0.862 & 0.644 \\
AP\_Omentum\_Uterus\_11 & 1.000 & 1.000 & 1.000 & 1.000 & 1.000 & 1.000 & 0.962 & 1.000 \\
OVA\_Colon\_1161 & 0.921 & 0.918 & 0.918 & 0.941 & 0.948 & 0.932 & 0.912 & 0.932 \\
OVA\_Omentum\_1139 & 0.991 & 0.986 & 0.986 & 0.982 & 0.960 & 0.975 & 0.711 & 0.944 \\
madelon\_1485 & 0.949 & 0.941 & 0.942 & 0.907 & 0.845 & 0.891 & 0.607 & 0.625 \\
\bottomrule
\end{tabular}
\end{adjustbox}
\end{table*}

\begin{table*}[h]
\centering
\caption{Healthcare \cite{bischl_openml_2021} Test Results (AUC $\uparrow$).\label{table:healthcare}}
\setlength{\tabcolsep}{2pt} % reduce horizontal padding
% \small % smaller font size
\begin{adjustbox}{width=1\textwidth} % scale table if still too wide
\begin{tabular}{ccccccccc}
\toprule
 \textbf{Dataset} & \textbf{TabPFN\textsubscript{n.e.} (RAD-TFM)} & \textbf{TabPFN\textsubscript{n.e.} (Base)} & \textbf{TabPFN\textsubscript{n.e.} (Finetuned)} & \textbf{CatBoost} & \textbf{RF} & \textbf{XGB} & \textbf{LR} & \textbf{MLP} \\
\midrule
SPECTF\_1600 & 0.844 & 0.852 & 0.852 & 0.844 & 0.825 & 0.809 & 0.780 & 0.728 \\
SPECTF\_337 & 0.949 & 0.935 & 0.936 & 0.950 & 0.951 & 0.924 & 0.928 & 0.717 \\
SPECT\_336 & 0.865 & 0.862 & 0.862 & 0.854 & 0.828 & 0.831 & 0.847 & 0.767 \\
abalone\_720 & 0.905 & 0.905 & 0.905 & 0.892 & 0.889 & 0.880 & 0.888 & 0.838 \\
audiology\_999 & 0.990 & 1.000 & 1.000 & 1.000 & 1.000 & 0.971 & 1.000 & 0.902 \\
blood-transfusion-se & 0.775 & 0.771 & 0.771 & 0.762 & 0.773 & 0.729 & 0.778 & 0.733 \\
bodyfat\_778 & 1.000 & 1.000 & 1.000 & 1.000 & 1.000 & 1.000 & 0.947 & 0.858 \\
breast-cancer-droppe & 0.538 & 0.532 & 0.532 & 0.450 & 0.421 & 0.404 & 0.503 & 0.456 \\
breast-cancer\_13 & 0.696 & 0.655 & 0.655 & 0.619 & 0.595 & 0.655 & 0.613 & 0.589 \\
cardiotocography\_146 & 1.000 & 1.000 & 1.000 & 1.000 & 1.000 & 1.000 & 0.978 & 1.000 \\
cardiotocography\_156 & 1.000 & 1.000 & 1.000 & 1.000 & 1.000 & 1.000 & 0.975 & 1.000 \\
chatfield\_4\_820 & 0.993 & 0.986 & 0.986 & 0.979 & 0.975 & 0.907 & 0.986 & 0.479 \\
dermatology\_1010 & 1.000 & 1.000 & 1.000 & 1.000 & 1.000 & 1.000 & 1.000 & 1.000 \\
haberman\_43 & 0.826 & 0.799 & 0.804 & 0.734 & 0.617 & 0.565 & 0.821 & 0.679 \\
heart-c\_982 & 0.924 & 0.933 & 0.933 & 0.916 & 0.891 & 0.887 & 0.920 & 0.882 \\
heart-h\_1565 & 0.608 & 0.606 & 0.607 & 0.705 & 0.694 & 0.659 & 0.541 & 0.641 \\
heart-h\_963 & 0.856 & 0.847 & 0.847 & 0.837 & 0.792 & 0.775 & 0.813 & 0.804 \\
mfeat-factors\_978 & 1.000 & 1.000 & 1.000 & 1.000 & 0.997 & 0.999 & 0.999 & 0.997 \\
mfeat-fourier\_971 & 1.000 & 1.000 & 1.000 & 1.000 & 1.000 & 1.000 & 1.000 & 1.000 \\
mfeat-karhunen\_1020 & 1.000 & 1.000 & 1.000 & 1.000 & 1.000 & 0.999 & 0.998 & 0.996 \\
mfeat-morphological\_ & 1.000 & 1.000 & 1.000 & 1.000 & 1.000 & 1.000 & 1.000 & 1.000 \\
mfeat-pixel\_1022 & 1.000 & 1.000 & 1.000 & 1.000 & 1.000 & 1.000 & 0.996 & 0.999 \\
mfeat-zernike\_995 & 1.000 & 0.999 & 0.999 & 1.000 & 0.999 & 1.000 & 0.972 & 0.968 \\
mu284\_880 & 1.000 & 1.000 & 1.000 & 1.000 & 1.000 & 1.000 & 1.000 & 0.738 \\
pbc\_810 & 0.753 & 0.757 & 0.757 & 0.778 & 0.765 & 0.831 & 0.705 & 0.703 \\
pbcseq\_802 & 0.833 & 0.857 & 0.855 & 0.902 & 0.901 & 0.933 & 0.728 & 0.666 \\
pwLinear\_721 & 0.950 & 0.960 & 0.960 & 0.980 & 0.980 & 1.000 & 0.970 & 0.700 \\
thoracic-surgery\_150 & 0.768 & 0.711 & 0.711 & 0.746 & 0.789 & 0.879 & 0.704 & 0.750 \\
thoracic\_surgery\_432 & 0.768 & 0.775 & 0.771 & 0.796 & 0.730 & 0.768 & 0.768 & 0.571 \\
vertebra-column\_1523 & 0.944 & 0.933 & 0.933 & 0.947 & 0.949 & 0.900 & 0.902 & 0.783 \\
vertebra-column\_1524 & 0.962 & 0.962 & 0.962 & 0.952 & 0.931 & 0.924 & 0.967 & 0.824 \\
\bottomrule
\end{tabular}
\end{adjustbox}
\end{table*}

%%%%%%%%%%%%%%%%%%%%%%%%%%%%%%%%%%%%%%%%%%%%%%%%%%%%%%%%%%%%
\clearpage
\newpage

\section{Additional Analyses}\label{app:additional-analyses}
\phantomsection\label{app:tabicl}\textbf{TabICL V2 Across Benchmarks.} Table \ref{table:tabicl-by-benchmark} reports the mean normalized AUC of the frozen, fine-tuned, and \model\-trained TabICL V2 classifier separately for each benchmark. In all three cases, the full method improves on both the original backbone and fine-tuning. These experiments use the same synthetic-generation and robust parameter-selection pipeline as TabPFN V2, without architecture-specific modifications. Normalization is performed over the expanded model comparison and therefore differs from the normalization used in Tables \ref{table:main} and \ref{table:main-apps}. Per-dataset AUC results for both TFM backbones and all five conventional baselines are reported in Tables \ref{table:tabpertnet-1}--\ref{table:pmlb-2}.
\begin{table}[h]
\centering
\caption{Mean normalized AUC for TabICL V2 variants on the three principal benchmarks.}
\setlength{\tabcolsep}{6pt}
\small
\begin{tabular}{lccc}
\toprule
\textbf{Benchmark} & \textbf{TabICL Base} & \textbf{TabICL FT} & \textbf{TabICL (\model)} \\
\midrule
PMLB & 0.892 & 0.911 & \textbf{0.924} \\
TabPertNet & 0.748 & 0.754 & \textbf{0.779} \\
TabArena & 0.891 & 0.894 & \textbf{0.914} \\
\bottomrule
\end{tabular}
\label{table:tabicl-by-benchmark}
\end{table}

\phantomsection\label{app:compute-matched}\textbf{Training-Budget Sensitivity.} To control for the number of training datasets, the fine-tuning runs use the same batch size and number of optimizer updates as the corresponding \model\ runs. The models are evaluated at checkpoints spaced by 500 training datasets. We report the mean test-AUC gap between each trained model and its frozen backbone, evaluated on the respective reference benchmark. In particular, continued fine-tuning on the real-data distribution degrades sharply beyond the earliest checkpoints, whereas the synthetic distribution selected by \model\ maintains positive gains throughout training (Table \ref{table:training-budget}). These comparisons control for dataset exposures and parameter-update counts, rather than total wall-clock time. The 30k \model\ entries correspond to selected checkpoints from training runs that processed 30k generated datasets in total, rather than evaluations at the terminal checkpoint. The fine-tuning entries at 30k report performance evaluated at that checkpoint.
\begin{table*}[h]
\centering
\caption{Mean raw test-AUC change relative to frozen TabPFN V2 at different training-dataset budgets. Positive values indicate improvement. At 30k, \model\ entries marked with an asterisk correspond to selected checkpoints from runs that processed 30k generated datasets, whereas fine-tuning entries are evaluated at the 30k checkpoint.}
\setlength{\tabcolsep}{4pt}
\small
\begin{tabular}{llrrrrrrr}
\toprule
\textbf{Benchmark} & \textbf{Method} & \textbf{500} & \textbf{1k} & \textbf{1.5k} & \textbf{2k} & \textbf{6k} & \textbf{10k} & \textbf{30k} \\
\midrule
TabArena & \model & +0.0005 & +0.0010 & +0.0016 & +0.0015 & +0.0026 & +0.0028 & $+0.0040^{*}$ \\
 & Fine-tuning & +0.0011 & +0.0015 & -0.0003 & -0.0024 & -0.0505 & -0.0840 & -0.0844 \\
\midrule
PMLB & \model & +0.0019 & +0.0025 & +0.0026 & +0.0033 & +0.0029 & +0.0030 & $+0.0030^{*}$ \\
 & Fine-tuning & -0.0006 & -0.0012 & -0.0017 & -0.0023 & -0.0166 & -0.0199 & -0.0233 \\
\midrule
TabPertNet & \model & +0.0031 & +0.0049 & +0.0052 & +0.0058 & +0.0083 & +0.0105 & $+0.0166^{*}$ \\
 & Fine-tuning & +0.0019 & +0.0009 & +0.0010 & +0.0001 & -0.0020 & -0.0097 & -0.0103 \\
\bottomrule
\end{tabular}
\label{table:training-budget}
\end{table*}

\phantomsection\label{app:components}\textbf{Individual and Combined Contributions.} To further quantify the role of the two components, we measure the raw-AUC gain over the frozen model from domain adaptation alone ($\Delta_{\mathrm{DA}}$), robustness alone ($\Delta_{\mathrm{R}}$), and their combination ($\Delta_{\mathrm{RAD}}$). The residual $\Delta_{\mathrm{RAD}}-\Delta_{\mathrm{DA}}-\Delta_{\mathrm{R}}$ captures the deviation from additivity, as shown in Table \ref{table:component-additivity}. The average gain under the full model ($+0.0055$) is close to the sum of the individual average gains ($+0.0052$); the corresponding residual is $+0.0003$. While the interaction varies across benchmarks, the two components contribute complementary improvements on average. Paired one-sided Wilcoxon signed-rank tests of normalized AUC over 236 datasets give $p=10^{-4}$ for domain adaptation alone, $p=0.028$ for robustness alone, and $p<10^{-4}$ for the combined method, each relative to the frozen model.
\begin{table*}[h]
\centering
\caption{Change in mean raw AUC from the frozen TabPFN V2 and the residual from additive contributions.}
\setlength{\tabcolsep}{6pt}
\small
\begin{tabular}{lrrrrr}
\toprule
\textbf{Benchmark} & $\boldsymbol{\Delta_{\mathrm{DA}}}$ & $\boldsymbol{\Delta_{\mathrm{R}}}$ & $\boldsymbol{\Delta_{\mathrm{DA}}+\Delta_{\mathrm{R}}}$ & $\boldsymbol{\Delta_{\mathrm{RAD}}}$ & \textbf{Residual} \\
\midrule
PMLB & +0.0033 & +0.0021 & +0.0053 & +0.0030 & -0.0023 \\
TabPertNet & +0.0074 & +0.0094 & +0.0167 & +0.0166 & -0.0001 \\
TabArena & +0.0013 & +0.0018 & +0.0031 & +0.0040 & +0.0010 \\
Genomics & +0.0024 & +0.0004 & +0.0028 & +0.0037 & +0.0009 \\
Healthcare & +0.0021 & +0.0023 & +0.0044 & +0.0035 & -0.0008 \\
Engineering & +0.0001 & -0.0016 & -0.0015 & +0.0020 & +0.0035 \\
\midrule
\textbf{Mean} & +0.0028 & +0.0024 & +0.0052 & +0.0055 & +0.0003 \\
\bottomrule
\end{tabular}
\label{table:component-additivity}
\end{table*}

\clearpage
\section*{NeurIPS Paper Checklist}

\begin{enumerate}

\item {\bf Claims}
    \item[] Question: Do the main claims made in the abstract and introduction accurately reflect the paper's contributions and scope?
    \item[] Answer: \answerYes{}
    \item[] Justification: The abstract and introduction state the paper's theoretical, algorithmic, and empirical contributions, and these are matched by the formal development in Sections \ref{sec:problem}--\ref{sec:rtfm} and the experiments in Section \ref{sec:experiments}.

\item {\bf Limitations}
    \item[] Question: Does the paper discuss the limitations of the work performed by the authors?
    \item[] Answer: \answerYes{}
    \item[] Justification: The paper discusses several concrete limitations, including limited statistical power for small benchmarks, challenging domains where the current SCM space and meta-features may be insufficient, and design restrictions such as fixed generated sample sizes; see Section \ref{sec:experiments} and Appendices \ref{app:da} and \ref{app:scm}.

\item {\bf Theory assumptions and proofs}
    \item[] Question: For each theoretical result, does the paper provide the full set of assumptions and a complete (and correct) proof?
    \item[] Answer: \answerYes{}
    \item[] Justification: The theoretical formulation and assumptions are introduced in Section \ref{sec:problem}, and formal proofs and supporting derivations are provided in Appendix \ref{app:proofs}.

	    \item {\bf Experimental result reproducibility}
	    \item[] Question: Does the paper fully disclose all the information needed to reproduce the main experimental results of the paper to the extent that it affects the main claims and/or conclusions of the paper (regardless of whether the code and data are provided or not)?
	    \item[] Answer: \answerYes{}
	    \item[] Justification: The paper specifies the algorithmic pipeline, benchmark construction, evaluation protocol, hyperparameters, and detailed pseudocode in Section \ref{sec:experiments} and Appendices \ref{app:da}, \ref{app:rtfm}, \ref{app:scm}, and \ref{app:experiments}.

\item {\bf Open access to data and code}
    \item[] Question: Does the paper provide open access to the data and code, with sufficient instructions to faithfully reproduce the main experimental results, as described in supplemental material?
    \item[] Answer: \answerNo{}
    \item[] Justification: We do not release code or additional assets with the current submission. The experiments rely on publicly available benchmarks and pretrained model backbones, and the method, optimization objective, parameter space, training protocol, and evaluation procedure are described in the paper and its appendices.

\item {\bf Experimental setting/details}
    \item[] Question: Does the paper specify all the training and test details (e.g., data splits, hyperparameters, how they were chosen, type of optimizer) necessary to understand the results?
    \item[] Answer: \answerYes{}
    \item[] Justification: Section \ref{sec:experiments} gives the main training setup and benchmark choices, and Appendices \ref{app:rtfm}, \ref{app:scm}, and \ref{app:experiments} provide additional details on splits, hyperparameters, and implementation choices.

\item {\bf Experiment statistical significance}
    \item[] Question: Does the paper report error bars suitably and correctly defined or other appropriate information about the statistical significance of the experiments?
    \item[] Answer: \answerYes{}
    \item[] Justification: Rather than error bars, we report formal statistical significance tests for the main claims, including Friedman tests and pairwise Wilcoxon signed-rank tests, as described in Section \ref{sec:experiments} and detailed further in Appendix \ref{app:experiments}. The tests span six benchmarks, with additional paired comparisons and bootstrap confidence intervals reported for the second TFM backbone and the component ablations.

\item {\bf Experiments compute resources}
    \item[] Question: For each experiment, does the paper provide sufficient information on the computer resources (type of compute workers, memory, time of execution) needed to reproduce the experiments?
    \item[] Answer: \answerYes{}
    \item[] Justification: The paper specifies the compute configuration and runtime breakdown used in practice, including GPU type and memory, CPU usage, domain-adaptation cost, and benchmark-dependent training duration; see Appendix \ref{app:experiments}.
    
\item {\bf Code of ethics}
    \item[] Question: Does the research conducted in the paper conform, in every respect, with the NeurIPS Code of Ethics \url{https://neurips.cc/public/EthicsGuidelines}?
    \item[] Answer: \answerYes{}
    \item[] Justification: To the best of our knowledge, this work conforms to the NeurIPS Code of Ethics. The paper studies generic tabular classification methods using cited benchmark resources and does not involve human subjects or the release of sensitive assets.

\item {\bf Broader impacts}
    \item[] Question: Does the paper discuss both potential positive societal impacts and negative societal impacts of the work performed?
    \item[] Answer: \answerNo{}
    \item[] Justification: The submission does not contain a dedicated broader-impact discussion. The work is methodological and application-agnostic, focusing on a generic tabular classifier rather than a specific downstream deployment setting.
    
\item {\bf Safeguards}
    \item[] Question: Does the paper describe safeguards that have been put in place for responsible release of data or models that have a high risk for misuse (e.g., pre-trained language models, image generators, or scraped datasets)?
    \item[] Answer: \answerNA{}
    \item[] Justification: The paper does not release high-risk generative models, scraped datasets, or similar assets that would require a safeguards discussion for controlled release.

\item {\bf Licenses for existing assets}
    \item[] Question: Are the creators or original owners of assets (e.g., code, data, models), used in the paper, properly credited and are the license and terms of use explicitly mentioned and properly respected?
    \item[] Answer: \answerYes{}
    \item[] Justification: We use existing benchmarks, datasets, software packages, and pretrained model weights. We cite the original creators in the main paper and provide an appendix paragraph listing the relevant licenses and terms of use for the principal assets, including TabArena, PMLB, OpenML, TabPertNet, Optuna, TabPFN-V2, and TabICL V2 in Appendix \ref{app:experiments}. We use these assets only for academic research and benchmarking and do not redistribute modified versions of the assets.

\item {\bf New assets}
    \item[] Question: Are new assets introduced in the paper well documented and is the documentation provided alongside the assets?
    \item[] Answer: \answerNA{}
    \item[] Justification: We do not release new code, models, datasets, or other assets as part of the current submission.

\item {\bf Crowdsourcing and research with human subjects}
    \item[] Question: For crowdsourcing experiments and research with human subjects, does the paper include the full text of instructions given to participants and screenshots, if applicable, as well as details about compensation (if any)? 
    \item[] Answer: \answerNA{}
    \item[] Justification: This work does not involve crowdsourcing or research with human subjects.

\item {\bf Institutional review board (IRB) approvals or equivalent for research with human subjects}
    \item[] Question: Does the paper describe potential risks incurred by study participants, whether such risks were disclosed to the subjects, and whether Institutional Review Board (IRB) approvals (or an equivalent approval/review based on the requirements of your country or institution) were obtained?
    \item[] Answer: \answerNA{}
    \item[] Justification: This work does not involve human subjects, so IRB approval and related participant-risk disclosures are not applicable.

\item {\bf Declaration of LLM usage}
    \item[] Question: Does the paper describe the usage of LLMs if it is an important, original, or non-standard component of the core methods in this research? Note that if the LLM is used only for writing, editing, or formatting purposes and does \emph{not} impact the core methodology, scientific rigor, or originality of the research, declaration is not required.
    %this research? 
    \item[] Answer: \answerNA{}
    \item[] Justification: LLMs are not an important, original, or non-standard component of the core methodology studied in this paper.

\end{enumerate}

\end{document}